\documentclass{article}

\usepackage{arxiv}

\usepackage{CJKutf8}

\usepackage[utf8]{inputenc} %
\usepackage[T1]{fontenc}    %
\usepackage{hyperref}       %
\usepackage{url}            %
\usepackage{booktabs}       %
\usepackage{amsfonts}       %
\usepackage{nicefrac}       %
\usepackage{microtype}      %
\usepackage{xcolor}         %
\usepackage{color}
\usepackage{wrapfig}
\usepackage{amsmath}
\usepackage{enumitem}
\usepackage{diagbox}
\usepackage{makecell}
\usepackage{multirow}
\PassOptionsToPackage{usenames, dvipsnames}{color}
\usepackage[usenames,dvipsnames]{color}
\usepackage{colortbl}
\definecolor{mygray}{gray}{.9}
\usepackage{array}
\usepackage{caption}

\usepackage{multibib}
\makeatletter
\@ifundefined{chapter}
  {
   }
  {
   }
\makeatother
\newcites{appendix}{References}

\newcommand{\methodname}{SIRLC}

\newcommand{\expect}{\mathbb{E}}

\def\gA{{\mathcal{A}}}

\def\gD{{\mathcal{D}}}

\def\gM{{\mathcal{M}}}

\def\gR{{\mathcal{R}}}
\def\gS{{\mathcal{S}}}

\usepackage{booktabs}       %
\usepackage{multirow}    
\usepackage{amsfonts}       %
\usepackage{nicefrac}       %
\usepackage{microtype}      %
\usepackage{natbib}

\usepackage{xcolor}

\usepackage{hyperref}

\usepackage{graphicx} %
\usepackage{float} %
\usepackage{subfigure} %
\usepackage{algorithm}
\usepackage{algorithmic}

\title{Language Model Self-improvement by Reinforcement Learning Contemplation}

\usepackage{authblk}

\author{%
  Jing-Cheng Pang\textsuperscript{\rm 1,2,*}, 
  Pengyuan Wang\textsuperscript{\rm 1,2,*}, 
  Kaiyuan Li\textsuperscript{\rm 1}, 
  Xiong-Hui Chen\textsuperscript{\rm 1,2}, 
  Jiacheng Xu\textsuperscript{\rm 1}, 
  Zongzhang Zhang\textsuperscript{\rm 1}, and 
  Yang Yu\textsuperscript{\rm 1,2,$\diamond$}\\
  \textsuperscript{\rm 1} National Key Laboratory for Novel Software Technology, Nanjing University, Nanjing, China \\
  \textsuperscript{\rm 2} Polixir.ai\\
  \textsuperscript{*} Equal contribution\\
  \textsuperscript{$\diamond$} Corresponding: yuy@nju.edu.cn
}

\date{}

\begin{document}

\maketitle
\begin{abstract}
    Large Language Models (LLMs) have exhibited remarkable performance across various natural language processing (NLP) tasks. However, fine-tuning these models often necessitates substantial supervision, which can be expensive and time-consuming to obtain. This paper introduces a novel unsupervised method called Language Model Self-Improvement by Reinforcement Learning Contemplation (\methodname) that improves LLMs without reliance on external labels. Our approach is grounded in the observation that it is simpler for language models to assess text quality than to generate text. Building on this insight, \methodname~assigns LLMs dual roles as both student and teacher. As a student, the LLM generates answers to unlabeled questions, while as a teacher, it evaluates the generated text and assigns scores accordingly. The model parameters are updated using reinforcement learning to maximize the evaluation score. We demonstrate that \methodname~can be applied to various NLP tasks, such as reasoning problems, text generation, and machine translation. Our experiments show that \methodname~effectively improves LLM performance without external supervision, resulting in a 5.6\% increase in answering accuracy for reasoning tasks and a rise in BERTScore from 0.82 to 0.86 for translation tasks. Furthermore, \methodname~can be applied to models of different sizes, showcasing its broad applicability.
\end{abstract}

\section{Introduction}
\label{sec:introduction}

Large language models (LLMs) have shown impressive performance in numerous natural language processing (NLP) tasks, including language understanding, machine translation, and question answering \cite{llm_survey,nlp_survey}. This success can be attributed to the Pre-training + Fine-tuning (PTFT) training framework, which involves training a language model on a large corpus and fine-tuning it on supervised NLP tasks. A fine-tuned language model can achieve state-of-the-art performance using various supervised datasets \cite{from_human_prefer}.
For example, InstructGPT \cite{instructgpt} and ChatGPT \cite{chatgpt} fine-tune the GPT-3 \cite{gpt3} model by introducing human preference and learning a reward model on human-comparison data.

However, fine-tuning LLMs typically requires extensive supervision in the form of labelled questions or human feedback, which can be time-consuming and labour-intensive. Recent research addresses this limitation by leveraging unlabelled data to improve LLMs' reasoning ability. For example, the self-consistency method \cite{self_consistency} samples diverse reasoning paths and selects the most consistent answer by marginalizing out the sampled paths. LMSI \cite{lm_self_improve} employs the self-consistency method to generate high-quality answers, which are then used to fine-tune LLMs. Although these methods improve performance using unlabelled data, they are primarily designed for reasoning tasks that rely heavily on LLMs' chain-of-thought (CoT) ability, which is limited to reasoning problems \cite{cot_prompt}. On the other hand, reinforcement learning shows an impressive performance in fine-tuning LLMs without directly using labelled answers \cite{instructgpt}, but it still requires amounts of annotation that reflects human preference and text quality.

\begin{wrapfigure}[21]{rt}{0.5 \textwidth}
\vspace{-1em}
\includegraphics[width=0.5 \textwidth]{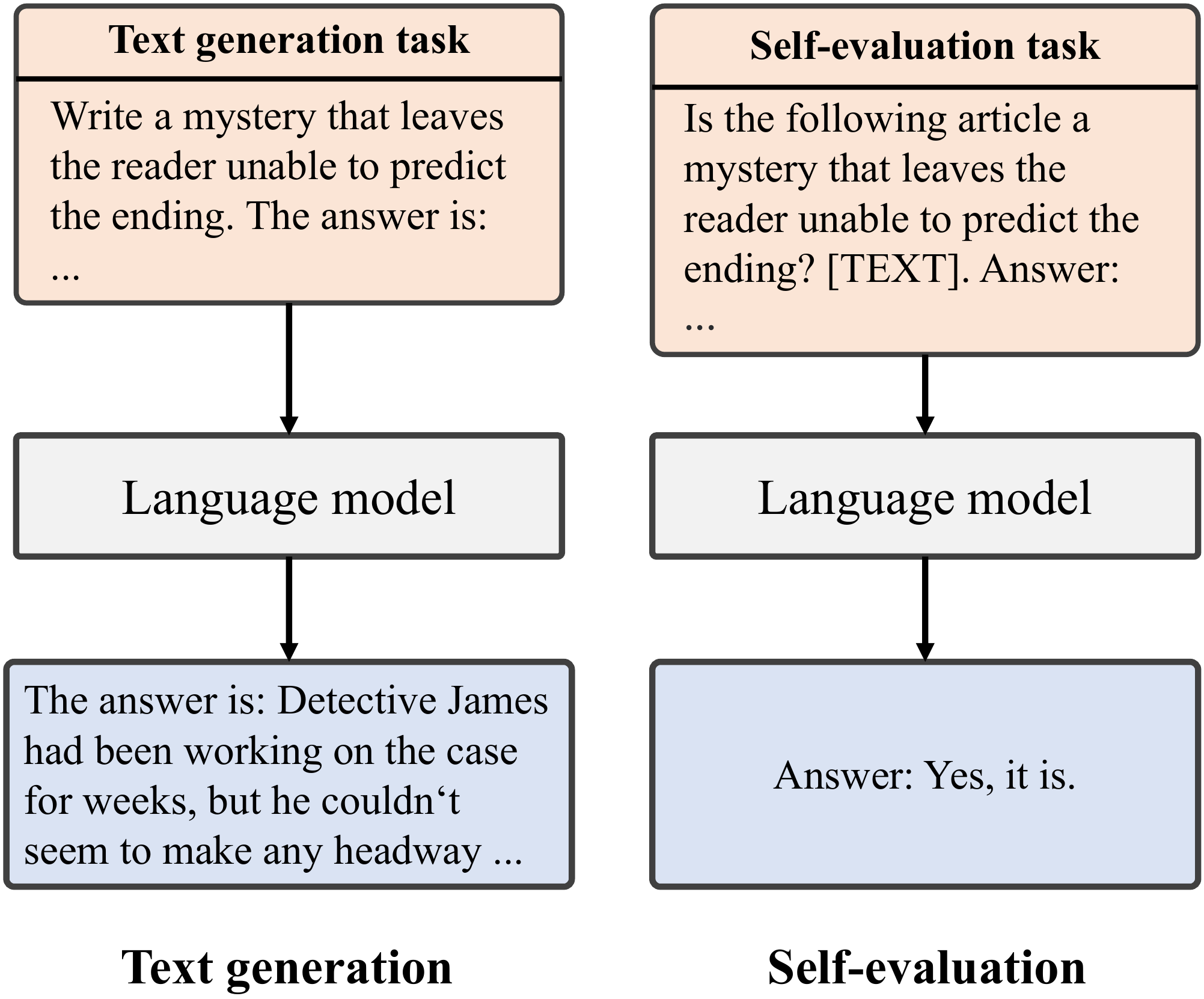}
\vspace{-0.85em}
\caption{A comparison between text generation and self-evaluation. Self-evaluation involves assessing and analyzing existing text, while generation requires the creation of entirely new text.}
\label{fig:generation_vs_evaluation}
\end{wrapfigure}

In this paper, we propose a novel approach for fine-tuning LLMs without external supervision. Our method capitalizes on the observation that it is simpler for a language model to evaluate the generated text than to generate it.
For example, while writing an attractive story can be challenging, identifying the generated text is relatively easy. Fig. \ref{fig:generation_vs_evaluation} illustrates the disparity between text generation and self-evaluation. 
We verify the self-evaluation ability of LLM through experiments on various NLP tasks. 
Based on such evaluation ability, we propose Language Model \textbf{S}elf-\textbf{I}mprovement by \textbf{R}einforcement \textbf{L}earning \textbf{C}ontemplation (\methodname), where the LLM both functions as a student and teacher. As a student, the LLM generates answers to unlabeled questions, while as a teacher, the LLM scores the generated answers. 
The LLM is subsequently updated through reinforcement learning to optimize for maximum evaluation scores. 
\methodname~employs self-evaluation results as the reward and utilizes reinforcement learning to retrain the LLM. We refer to this learning process as \emph{reinforcement learning contemplation}.

The contribution of this work can be summarized as follow:
Firstly, We introduce a novel approach for unsupervised fine-tuning of LLMs by utilizing self-evaluation as the reward and RL for training, eliminating the need for external supervision. Secondly, we conduct a comprehensive experimental analysis to demonstrate LLM's self-evaluation ability. To the best of our knowledge, this is the first study that formally verifies the self-evaluation capability of LLMs. Finally, our experimental results demonstrate that our approach can improve LLM's ability to solve reasoning, summarization, and translation problems. We also present that \methodname~can be applied to LLMs with a parameter range of 80M to 780M, and the trained LLM generalizes well to new and unseen datasets, demonstrating the extensive applicability of the proposed method.

\section{Related Work}
\label{sec:related_work}

\textbf{Train language model with unlabeled data.}
Learning from unlabelled data is a promising approach that eliminates the need for extensive annotation. Self-training is a popular technique in this field, which entails assigning pseudo labels from a learned classifier to unlabeled data. These pseudo-labelled examples are then utilized to enhance the initial model training \cite{self_training_1,self_training_2,self_training_3}. In recent years \cite{lm_self_improve}, self-training has been employed to fine-tune large-scale language models such as PaLM \cite{palm} with 540B parameters, and it has shown immense potential. However, this approach heavily relies on the CoT prompt, which is limited to solving reasoning problems. In contrast to methods that train LLMs to fit pseudo-labels, our approach employs reinforcement learning to train LLMs, which has proven more effective than supervised learning \cite{rl_train_llm}.

\textbf{Train language models with reinforcement learning.} 
RL has demonstrated significant success in training language models \cite{rl_train_llm,rl_train_llm_2}. For instance, some studies train LLMs by utilizing algorithmically defined reward functions for specific NLP tasks, such as BLEU for translation \cite{train_with_bleu,rl_llm_related_1} and ROUGE for summarization \cite{train_with_rouge}. In a departure from the heuristic definition of reward functions, another approach \cite{rl_llm_related_2} involves learning to evaluate text coherence and utilizing the learned model to provide rewards for summarization and long-form generation tasks.
Incorporating human preferences into language models using reinforcement learning has recently gained popularity. A series of works \cite{from_human_prefer,instructgpt,chatgpt} develop reward models reflecting human preferences and train language models using RL algorithms. However, all these related studies necessitate a pre-defined reward function or a reward model learned from annotated data.
In contrast, our method eliminates the need for external labels or reward models. Instead, the language model is updated to maximize self-evaluation scores.

\textbf{Self-evaluation of LLM.} 
Prior research has effectively utilized the self-evaluation ability of LLM to identify errors in previously generated text \cite{lm_self_improve,self_evaluation_revise_mistakes,self_evaluation_self_verification,self_evaluation_re_prompt,self_consistency}. For instance, the re-prompting method \cite{self_evaluation_re_prompt} detects errors in the current plan generated by LLM and revises the current step if an error is discovered. Self-verification \cite{self_evaluation_self_verification} assesses the accuracy of the generated answer by using it as a condition to construct a new task, subsequently prompting the LM to re-predict the original masked conditions. Self-consistency \cite{self_consistency} implicitly employs the self-evaluation ability of LLM by selecting the most consistent answer from a diverse set of reasoning paths.
Additionally, some studies have used LLM to score the generated text directly. For example, the generate \& rank method \cite{generate_and_rank} trains a language model to rank answers based on a scoring system. \cite{from_human_prefer} develops a smaller language model as a reward model to evaluate the alignment between generated text and human-produced text. While these previous works leverage self-evaluation ability (to be more exact, only evaluation ability in some works) to rectify generation errors, our study showcases the self-evaluation ability through experimental results. It directly employs self-evaluation to determine the accuracy and quality of the text.

\section{Preliminary}

\label{sec:preliminary}
We begin with a vocabulary $\Sigma$ and an LLM $\gM$ which takes a token sequence of the question $q = \{q_0, ..., q_n\}$ as input and predicts the next token using autoregressive modelling: $o_{t+1} = \gM(q, \{o_0, \dots, o_t\})$, where $q\in\Sigma^n$ and $o_t \in \Sigma$.
To fine-tune LLM with RL, we can view this problem as a Markov Decision Process (MDP) \cite{puterman2014markov, sutton2011reinforcement}, which is described as a tuple $\left( \gS, \gA, P, \gR, \gamma, d_0 \right)$:
\begin{itemize}
    \item State space $\gS$: the space of input token sequences $q \cup \{o_0,o_1,\dots,o_t\}$.
    \item Action space $\gA$: the space of tokens $o_t$.
    \item Reward function $\gR(q, \{o_0,\dots, o_t\})$: a score that reflects the quality of the generated answer to the question, which can be obtained from human feedback or a predefined criterion. The reward is typically given when the complete answer has been generated.
    \item Transition $P$: $s_{t+1}=s_t \cup o_{t+1}$.
    \item Initial distribution $d_0$: the distribution of question $q$.
\end{itemize}

Here, the LM $\gM$ acts as a policy mapping from state space to the probability space over action space. 
The objective of RL is to train the policy to maximize the expected returns:
\begin{equation}
    \expect \bigg[ \sum_{t=0}^\infty \gamma^t \gR(q,o) \big|q \sim d_0, o_t \sim \gM (\cdot|s_t)\bigg].
\end{equation}

In \methodname, we formulate the LLM fine-tuning problem as an MDP, analogous to the one described above, with the reward function derived from self-evaluation results. To simplify the notation, we use $o \sim \gM(q)$ to represent the autoregressive sampling of a complete answer $o$ from the language model $\gM$, based on the input question $q$.

\section{Large Language Models are Good at \\ Self-evaluation}
\label{sec_evaluation_ability_ver}

In this section, we aim to verify the self-evaluation ability of LLMs by investigating three key topics in subsequent subsections: (1) comparison of LLMs' ability in text generation and self-evaluation; (2) the correlation between self-evaluation results and the established evaluation metrics; and (3) potential for self-evaluation to improve LLMs.

\subsection{Comparison of Text Generation and Self-evaluation}
\label{sec:comparison_of_tg_se}
We conduct experiments to compare the text generation and self-evaluation abilities of LLMs using the CommonGen \cite{commongen} task, which involves generating a sentence that describes an everyday scenario based on a given set of common concepts such as \emph{\{dog, frisbee, catch, and throw\}}. Specifically, we use FLAN-T5 \cite{flan-t5} as the LLM to generate text based on common conceptual questions provided by CommonGen and evaluate the accuracy of the generated text through human evaluation. In addition, we assess the LLM's text evaluation ability by using the same LLM to evaluate whether the generated text meets the given concepts. Appendix \ref{appendix:exp_detail} shows more experiment details (e.g., the prompts we use). As the experiment results presented in Fig. \ref{fig:se_vs_tg}, we observe that the self-evaluation accuracy exceeds the generation accuracy in all scales of models. Especially when the parameter size of the model is small (see FLAN-T5-Large/XL), the self-evaluation accuracy significantly outperforms the text generation by 15\%. These results indicate that it is simpler to self-evaluate the generated text than to generate high-quality text that meets contextual requirements. However, it is essential to note that the evaluation accuracy is affected by the quality of the generated text, and this experiment is only a preliminary demonstration of the LLM's ability to self-evaluate. We conduct more experiments to further verify the self-evaluation ability, as presented in the following subsections.

\begin{table}[htbp]
\centering
  \begin{minipage}[t]{0.35\linewidth}
    \centering
    \vspace{-3.5em}
    \subfigure{
    \includegraphics[width=1\textwidth]{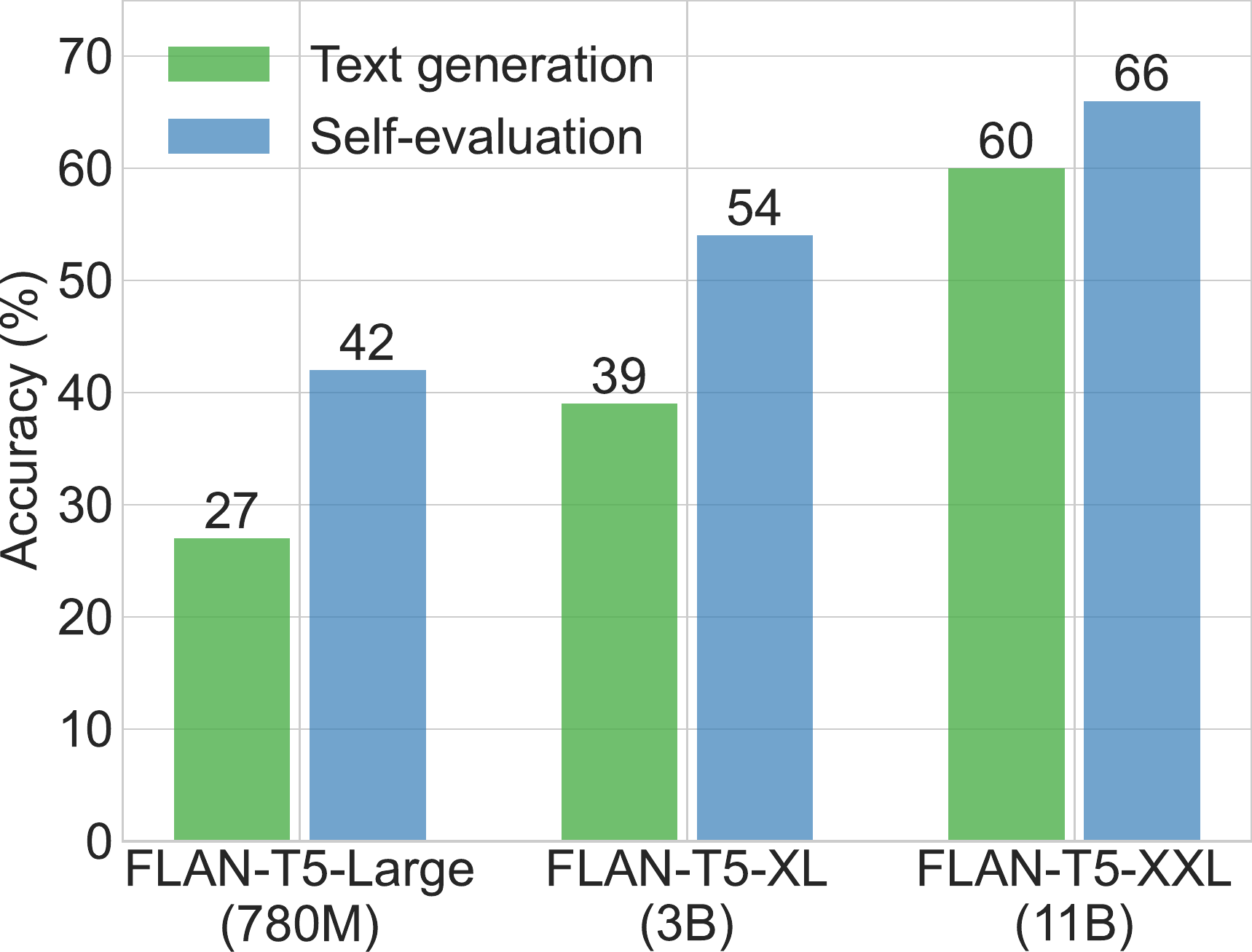}
    }
    \captionof{figure}{Comparison of the text generation and self-evaluation.} 
    \label{fig:se_vs_tg}
  \end{minipage} \hspace{0.6em}
  \begin{minipage}[t]{0.6\linewidth}
    \centering
    \small
    \begin{tabular}{c|ccc}
       \toprule
       \diagbox{Task}{Met.} & BLEU & ROUGE & BERTScore \\ \midrule
       CNN & 0.26 & 0.16 & 0.23 \\
       IWSLT 2017 & 0.21 & 0.28 & 0.29 \\
       \bottomrule
    \end{tabular}
    \vspace{1em}
    \caption{The correlation coefficient between self-evaluation and different metrics. The self-evaluation result correlates positively with all three metrics. The experiments are conducted with FLAN-T5-XL.}
    \label{tab:self_evaluation_vs_metrics}
  \end{minipage}
  \vspace{-1em}
\end{table}

\subsection{Correlation Between Self-evaluation and Established Metrics}
\label{sec:correlation_se_em}
This section provides an analysis of the correlation between self-evaluation and established metrics using two datasets: the CNN/Daily Mail dataset \cite{cnn_daily_mail} for text summarization and the IWSLT 2017 dataset \cite{iwslt_2017} for text translation. To find the relationship between self-evaluation and established metrics, LLM generates two answers for each question in these tasks and then self-evaluates to determine which answer is better. Additionally, we utilize three metrics, namely BLEU \cite{bleu_score}, ROUGE \cite{rough_score}, and BERTScore \cite{bert_score}, to compare the two answers, calculate their correlation coefficient with self-evaluation, and present the findings in Tab. \ref{tab:self_evaluation_vs_metrics}. As demonstrated by the results, the self-evaluation exhibits a positive correlation with all the considered metrics, indicating a consistent measurement of the quality of the generated text, and it is a reliable way to measure the quality of the generated text.

\subsection{Potential for Self-improvement}
\label{sec:potential_for_self_inp}
The previous section shows a positive correlation between self-evaluation and established metrics. However, it remains to be seen whether self-evaluation can be utilized to improve text generation. To investigate this problem, we design a text generation strategy that re-samples the answer based on the self-evaluation result, which will be explained in the following paragraphs. In order to evaluate the effectiveness of this approach, we conduct experiments on various NLP benchmarks.

One such benchmark is Bigbench-hard \cite{bigbench}, which includes multiple reasoning tasks consisting of multiple-choice and direct-generation tasks. We compare two answer generation strategies: (1) \textbf{w/o SE}: the answer is a directly deterministic output of the LLM, and (2) \textbf{w/ SE}: LLM generates an answer and evaluates its correctness. If the evaluation indicates the answer is incorrect, LLM re-generates an answer as the final output. We tested these two strategies on multiple Bigbench-hard tasks, and the results are presented in Tab. \ref{tab:non-invasive-self-improve}. The experiment results demonstrate that with self-evaluation, the answer accuracy outperforms that of direct answer generation on 11 of 12 evaluation tasks. This result justifies using self-evaluation to help LLM improve answer accuracy.

\begin{table}[htbp]
\setlength{\tabcolsep}{0.5mm}
\small
    \centering
    \begin{tabular}{c|cccc}
    \toprule
   & \makecell[c]{Reasoning about \\ Colored Objects} & \makecell[c]{Logical \\ Deduction (7)} & \makecell[c]{Tracking Shuffled \\ Objects (5)} & \makecell[c]{Object \\ Counting}  \\
   \midrule
   w/o SE & 30.9\% & 18.5\% & 10.1\% & 34.7\%  \\
   w/ SE & \textbf{31.1\%} & \textbf{20.5\%} & \textbf{11.1\%} & \textbf{34.9\%}  \\
   \bottomrule
   \toprule
   & \makecell[c]{Web of Lies} & \makecell[c]{Sports \\ Understanding} & \makecell[c]{Logical \\ Deduction (3)} & \makecell[c]{Logical \\ Deduction (5)}  \\
   \midrule
    w/o SE & 51.6\% & 59.7\% & 34.9\% & 23.6\%   \\ 
    w/ SE & \textbf{53.2\%} & 59.7\% & \textbf{38.3\%} & \textbf{25.7\%}  \\
   \bottomrule
   \toprule
   & \makecell[c]{Penguins in \\ a Table} & \makecell[c]{Navigate} & \makecell[c]{Tracking Shuffled \\ Objects (3)} & \makecell[c]{Geometric \\ Shapes} \\
   \midrule
    w/o SE &  23.5\% & 47.7\% & 28.1\% & 10.7\% \\ 
    w/ SE & \textbf{28.8\%} & \textbf{50.5\%} & \textbf{31.5\%} & \textbf{13.5\%} \\
   \bottomrule
\end{tabular}
    \vspace{1em}
    \caption{Comparison of the answer accuracy between answer generation with/without self-evaluation. Full results on all 27 BigBench tasks are presented in Appendix \ref{appendix:add_exp_full_self_evaluation}.}
    \label{tab:non-invasive-self-improve}
\end{table}

\begin{figure}[htbp]
    \centering
    \subfigure[BLEU]{
        \includegraphics[width=0.3\textwidth]{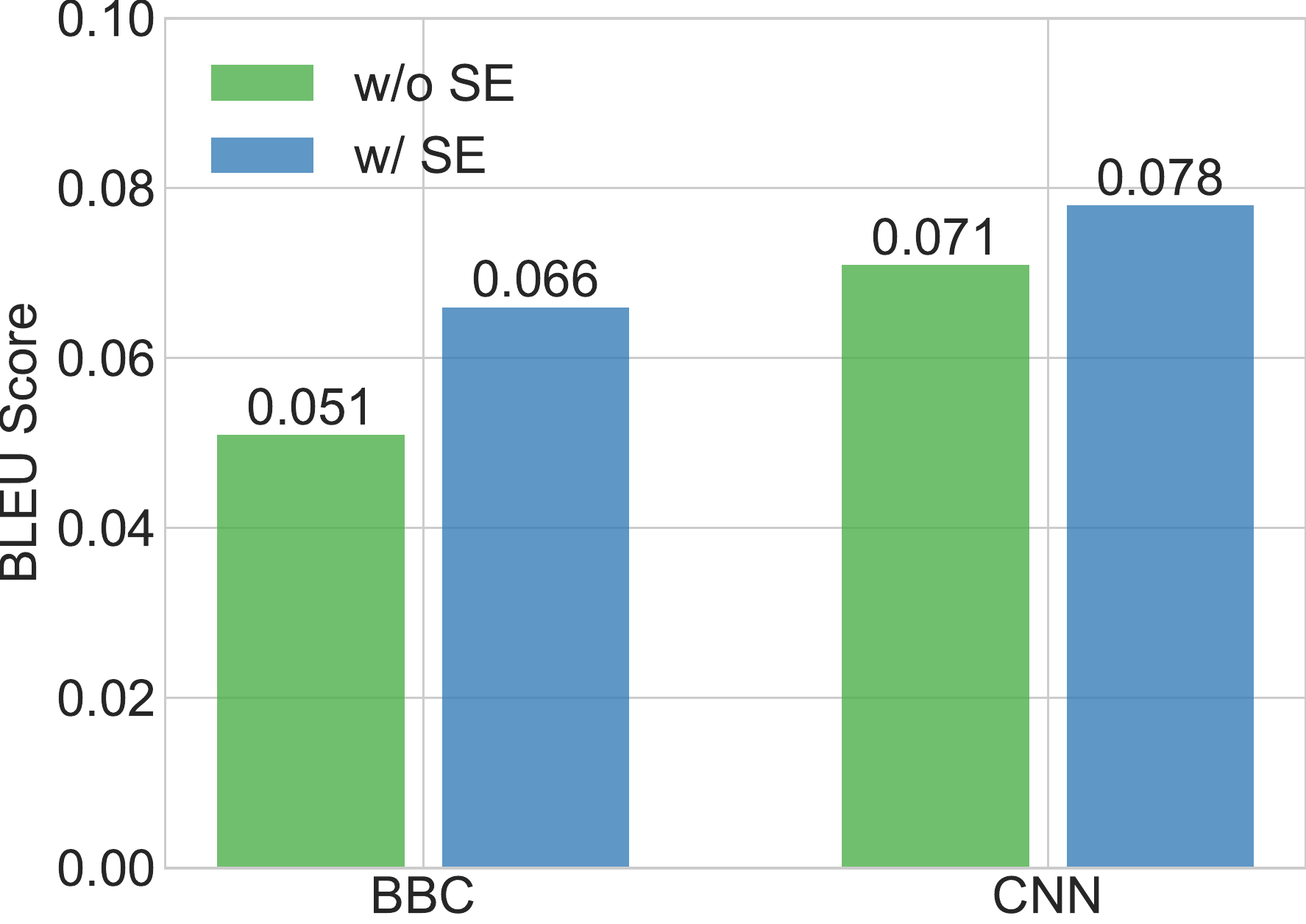}
    }
    \subfigure[ROUGE]{
        \includegraphics[width=0.3\textwidth]{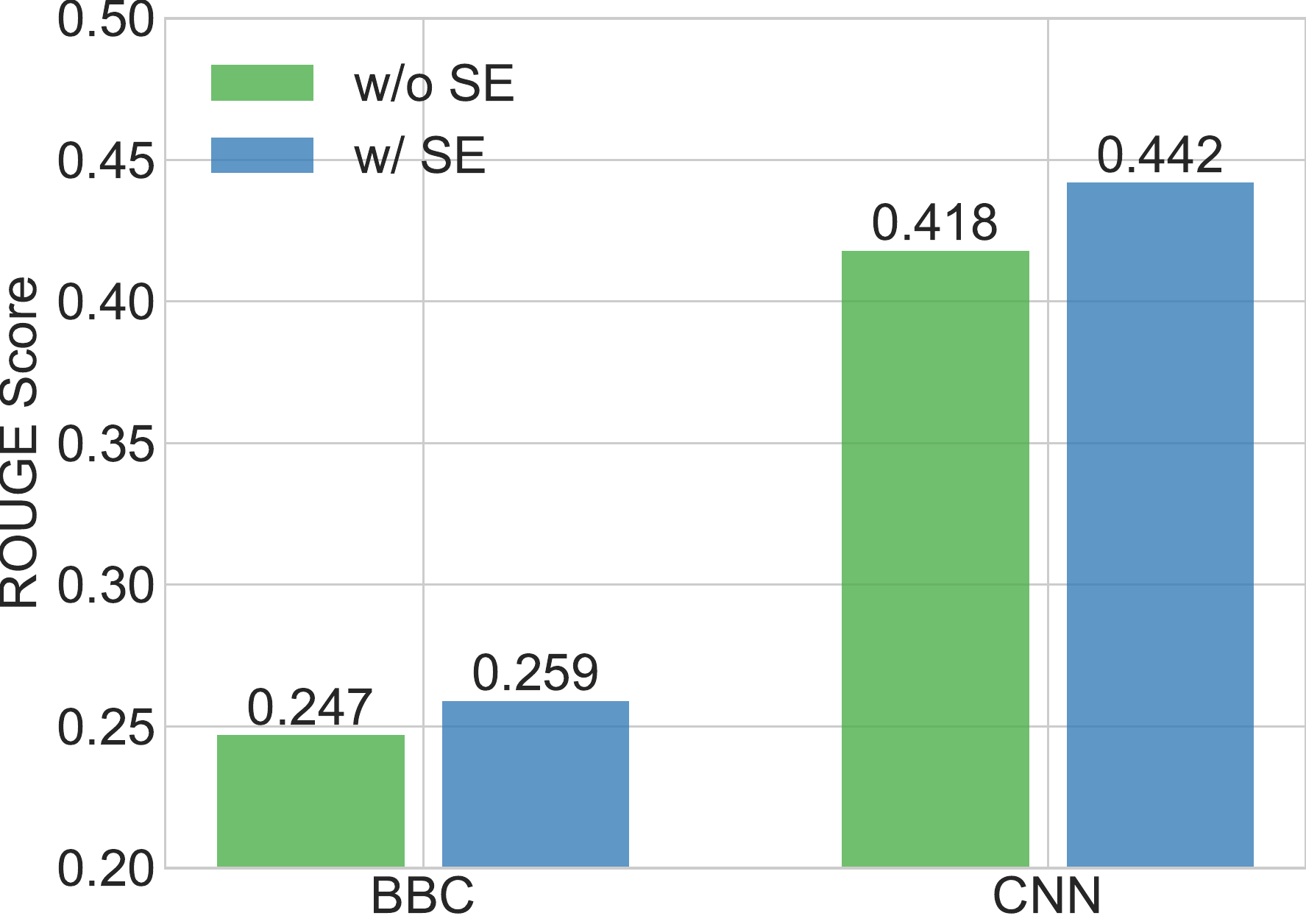}
    }
    \subfigure[BERTScore]{
        \includegraphics[width=0.3\textwidth]{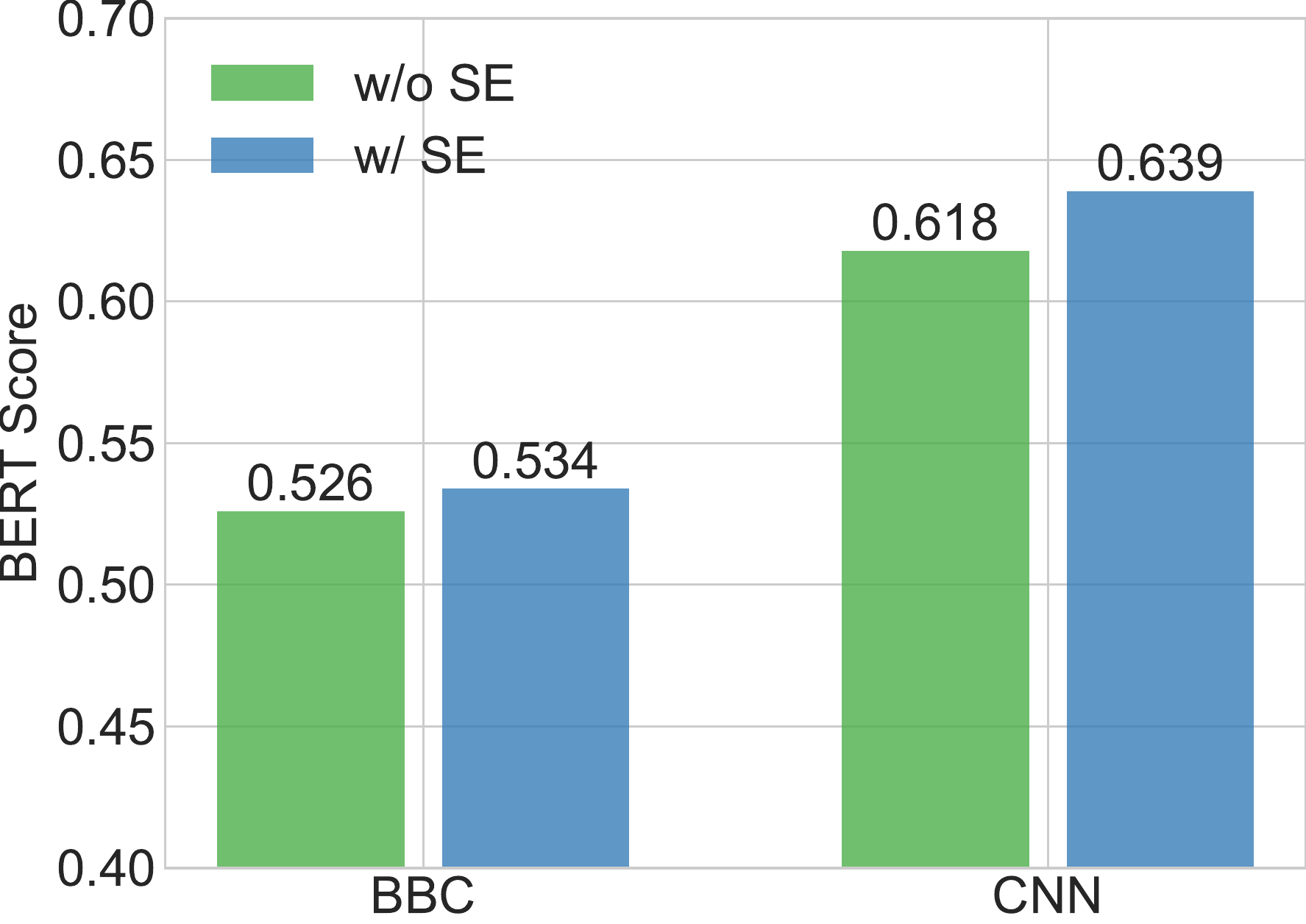}
    }
    \caption{Comparison of text generation with/without self-evaluation on text summarization tasks.}
    \label{fig:with_and_without_se_on_summary}
\end{figure}

Furthermore, we also conduct experiments on two text summarization tasks, CNN/Daily Mail and BBC \cite{bbc_dataset}. As it is not meaningful to evaluate the correctness of generated summarizations, we use a different approach to utilize self-evaluation in this experiment: (1) \textbf{w/ SE}: LLM samples three different answers and evaluates which answer is the best one as the final answer, and (2) \textbf{w/o SE}: LLM samples three different answers, and we present the average score of the three answers. As shown in Fig. \ref{fig:with_and_without_se_on_summary}, the generated answers have higher scores under all three metrics when self-evaluation is used. This result suggests that self-evaluation can potentially improve the quality of the generated text, which serves as a stepping stone for building \methodname~method.

\section{Self-improvement by Reinforcement Learning Contemplation}

\label{sec:method}

\begin{figure*}[t]
\centering
\includegraphics[width=1\linewidth]{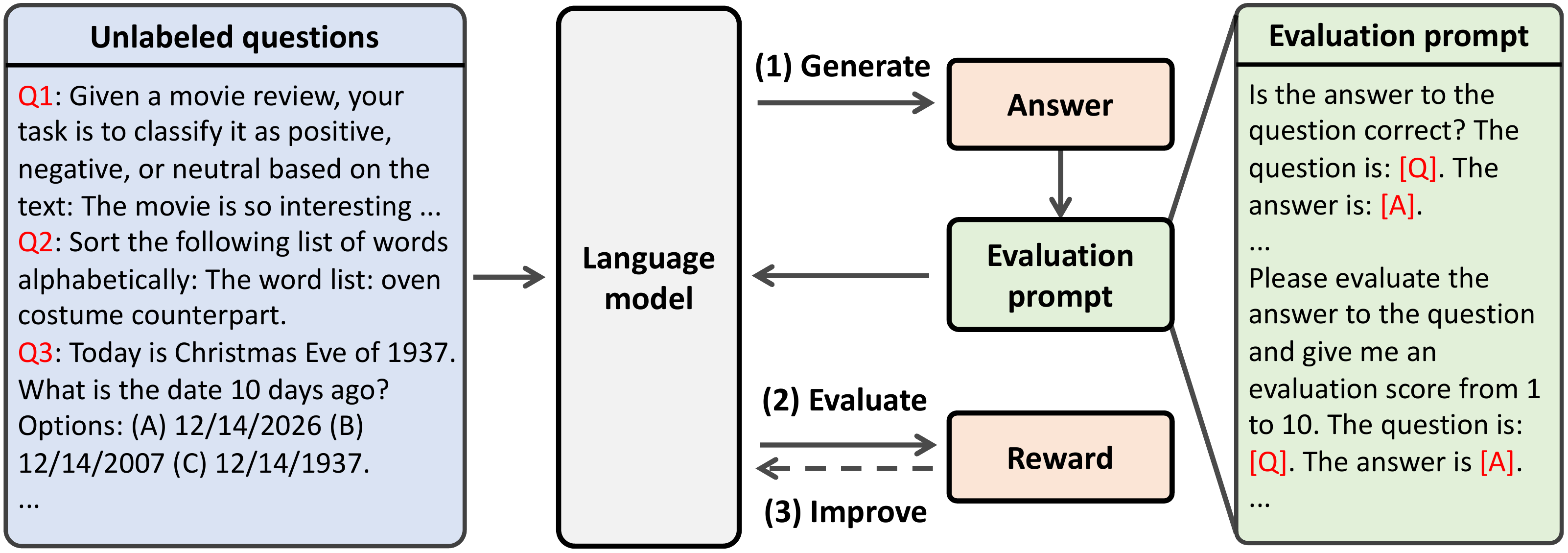}
\caption{Overall training procedure of \methodname, which iterates through three steps: (1) Answer generation to the unlabeled questions. (2) Self-evaluation by asking LM using \textit{evaluation prompt}, with the evaluation results as the reward. (3) Update the language model to maximize the reward using reinforcement learning algorithms. The solid lines represent the data flow, while the dashed line represents the update of LLM parameters.
}
\label{fig:overall_framework}
\end{figure*}

In the previous section, we observe that self-evaluation has the potential to be a helpful tool for LLM self-improvement. 
In this section, we will elaborate on our method, \methodname, that improves LLM ability without external labels based on self-evaluation.
We are given a pre-trained language model $\mathcal{M}$ and an unlabeled training dataset $\gD^\text{train} = \{q_i\}_{i=1}^{|\gD|}$. 
The overview of our method is illustrated in Fig. \ref{fig:overall_framework}, which iterates through the following steps:
\begin{itemize}
    \item Gathering question-answer pair $(q, o)$ via $q \sim \gD^\text{train}$, $o \sim \gM$.
    \item Self-evaluation on the question-answer, and obtain the reward $r$.
    \item Self-improvement with reinforcement learning.
\end{itemize}

\textbf{Answer generation to unlabeled questions.}
We randomly sample a batch of questions from $\gD^\text{train}$ and ask the LLM to generate answers to these questions with a temperature of $T=1$. We use different prompts to generate answers for different questions, as described in Appendix \ref{appendix:exp_prompts}. For reasoning problems, we use the Chain-of-Thought (CoT) prompt, which has been shown to improve LLM performance in previous studies \cite{self_consistency,lm_self_improve}.

\textbf{Self-evaluation as the reward.} 
After gathering the question-answer pair $(q,o)$, \methodname~employs the LLM to evaluate the reward of the generated answer:
\begin{equation}
    R(q,o) = \phi(\gM(p_\text{EP},q,o)),
    \label{eq:reward}
\end{equation}
where $\phi$ is a text processing function that processes the LLM output to the numerical reward and $p_\text{EP}, $ is the prompt used for self-evaluation. 
\methodname~uses two types of evaluation prompts: (1) Correctness Evaluation Prompt (\textbf{CEP}): \emph{"Is the answer to the question correct? The question is: [Q]. The answer is: [A]"}, and (2) Quality Evaluation Prompt (\textbf{QEP}): \emph{"Please evaluate the answer to the question and give me an evaluation score from 1 to 10. The question is: [Q]. The answer is [A]"}. Depending on the type of question, either CEP or QEP is utilized to evaluate the generated text.

\methodname~applies CEP to assess the factual accuracy of the generated text, which is essential for tasks such as question-answering and reasoning. The CEP prompts LLMs to verify the answer's correctness and identify any factual errors. Given a question $q$ and the answer $o$ generated by the LLM, the reward $R(q,o)$ is a 0-1 value based on the evaluation result. \methodname~employs QEP to evaluate the overall effectiveness and coherence of the generated text in terms of its writing, structure, and style. Quality is often assessed on a scale, with higher scores indicating better overall effectiveness and coherence of the text. The QEP prompts LLMs to rate the text's quality on a scale of 1 to 10 based on how well it answers the question. This prompt type is helpful for text-generation tasks such as translation and summarization. 

During training, the reward distribution may change as the LLM is updated. We use the initial pre-trained LLM $\gM^*$ for self-evaluation while keeping it fixed to ensure stable training.

\textbf{Self-improvement through reinforcement learning.} 
With the evaluation reward, the LLM $\gM$ can be updated using any RL algorithm to maximize the reward. In our case, we employ the Proximal Policy Optimization (PPO) algorithm \cite{PPO}, which has demonstrated promising results in applications \cite{ppo_application}. To ensure better exploration, we apply entropy regularization, which prevents the sampling phase from converging too early. Additionally, we use the Kullback-Leibler (KL) divergence to prevent $\gM$ from deviating too far from the initial pre-trained LLM.
\section{Experiment}
\label{sec:exp}

We conduct a series of experiments to demonstrate the effectiveness of our proposed \methodname~method. 
Our experiments encompass the following topics: (1) comparison of the proposed method with baseline methods on various NLP tasks (Section \ref{sec:exp_main_res}); (2) the generalization ability of the LLM on unseen datasets after fine-tuning using the proposed method (Section \ref{sec:exp_perf_of_met}); and (3) the application of the proposed method to different sizes of language models (Section \ref{sec:exp_perf_of_met}). We first introduce our experimental setup in the subsequent subsection.

\subsection{Experiment Setup}

\textbf{Dataset for evaluation.} We consider various NLP tasks that focus on the different abilities of LLM, which can be divided into three categories: (1) \textbf{Reasoning problem}: BigBench \cite{bigbench} is a challenging generation task that requires complex reasoning capabilities of the language models. The tasks in BigBench are pretty diverse, including reasoning the final results of a sequence of actions, understanding dates, and completing tasks that require simple arithmetic calculations. In our experiments, we use 12 challenging tasks \footnote{Detailed descriptions about the tasks are in \url{https://github.com/google/BIG-bench/blob/main}.}, which covers multiple-choices, judgments and text generation tasks. 
(2) \textbf{Language translation}: IWSLT 2017 \cite{iwslt_2017} dataset includes data in a variety of languages, including English, German, French, Chinese, Japanese, and Arabic, which has been widely used in machine translation research. 
(3) \textbf{Text summarization}: CNN/Daily Mail \cite{cnn_daily_mail} and BBC \cite{bbc_dataset} are two popular datasets used for text summarization tasks. The CNN/Daily Mail dataset covers a wider range of topics than the BBC dataset, including politics, sports, and entertainment. In contrast, the BBC dataset focuses more on news and current events. See Tab. \ref{tab:example_of_dataset} for examples of the tasks used in our experiments.

\begin{table}[htbp]
\small
    \centering
    \begin{tabular}{c|p{5.5cm}|p{3cm}}
   \toprule
   & \centering Example inputs & Example outputs \\
   \midrule
   Judgement &  ``Lionel Messi hit a three-run homer. Answer (`Plausible' or `Implausible').'' & "Implausible"\\ 
   Text generation &  “I have a flute, a piano, a trombone, four stoves, a violin, an accordion, a clarinet, a drum, two lamps, and a trumpet. How many musical instruments do I have?” & “8” \\ \midrule
   CNN/Daily & \centering ``Summarize the following article: The National Football League has indefinitely suspended Atlanta Falcons quarterback ...'' & ``The NFL has suspended Atlanta Falcons quarterback Michael Vick ...'' \\ \midrule
   ITSLW 2017 & \centering ``Please translate the following Chinese text into English. Text:
   \begin{CJK*}{UTF8}{gbsn} 你好，世界.''\end{CJK*} & ``Hello, world.'' \\ 
   \bottomrule
\end{tabular}
    \vspace{1em}
    \caption{Examples of inputs and outputs for the tasks in our experiments.}
    \label{tab:example_of_dataset}
\end{table}

\textbf{Baselines for comparison.} 
We compare \methodname~with representative methods that improve LLM without supervised data. These methods include:
(1) Self-consistency (\textbf{SC}) \cite{self_consistency} samples a diverse set of reasoning paths instead of solely relying on the greedy LLM output. It then selects the most consistent answer by marginalizing out the sampled reasoning paths. In our experiments, we use the number of sample paths as three. 
SC is typically applicable to reasoning problems, as it involves voting among different reasoning paths. 
(2) \textbf{LMSI} \cite{lm_self_improve} utilizes the SC method to generate "high-confidence" answers for unlabeled questions. It then fine-tunes the LLM using the self-generated solutions as target outputs.
Additionally, we consider the following:
(3) Reinforcement Learning Fine-Tuning (\textbf{RLFT}) fine-tunes the LLM using reinforcement learning (RL) and employs oracle metrics as the reward. The metric used is answer accuracy for tasks such as multiple-choice, judgment, and fill-in-the-blank. For translation and summarization tasks, BERTScore is used as the metric.
(4) \textbf{DG} directly generates the answer using the deterministic output of the LLM.

\textbf{Implementation details.} 
We employ PPO to train the LLM for 6,000 gradient steps for each task, with each batch size of 12. We utilize the trlx repository from GitHub \cite{trlx} to implement PPO. We implement SIRLC using CEP in reasoning tasks while employing QEP for other tasks. Unless otherwise specified, we use FLAN-T5-Large, which has 780M parameters, as our LLM in the experiments. All reported results are averaged over three random trials, and the experiments can be conducted using two GTX 3090 graphics cards with 24GB of memory. We provide specific hyperparameters and more detailed implementation descriptions in Appendix \ref{appendix:exp_detail}.

\begin{figure}[t]
    \centering
    \subfigure[Geometric Shapes]{
        \includegraphics[width=0.3\textwidth]{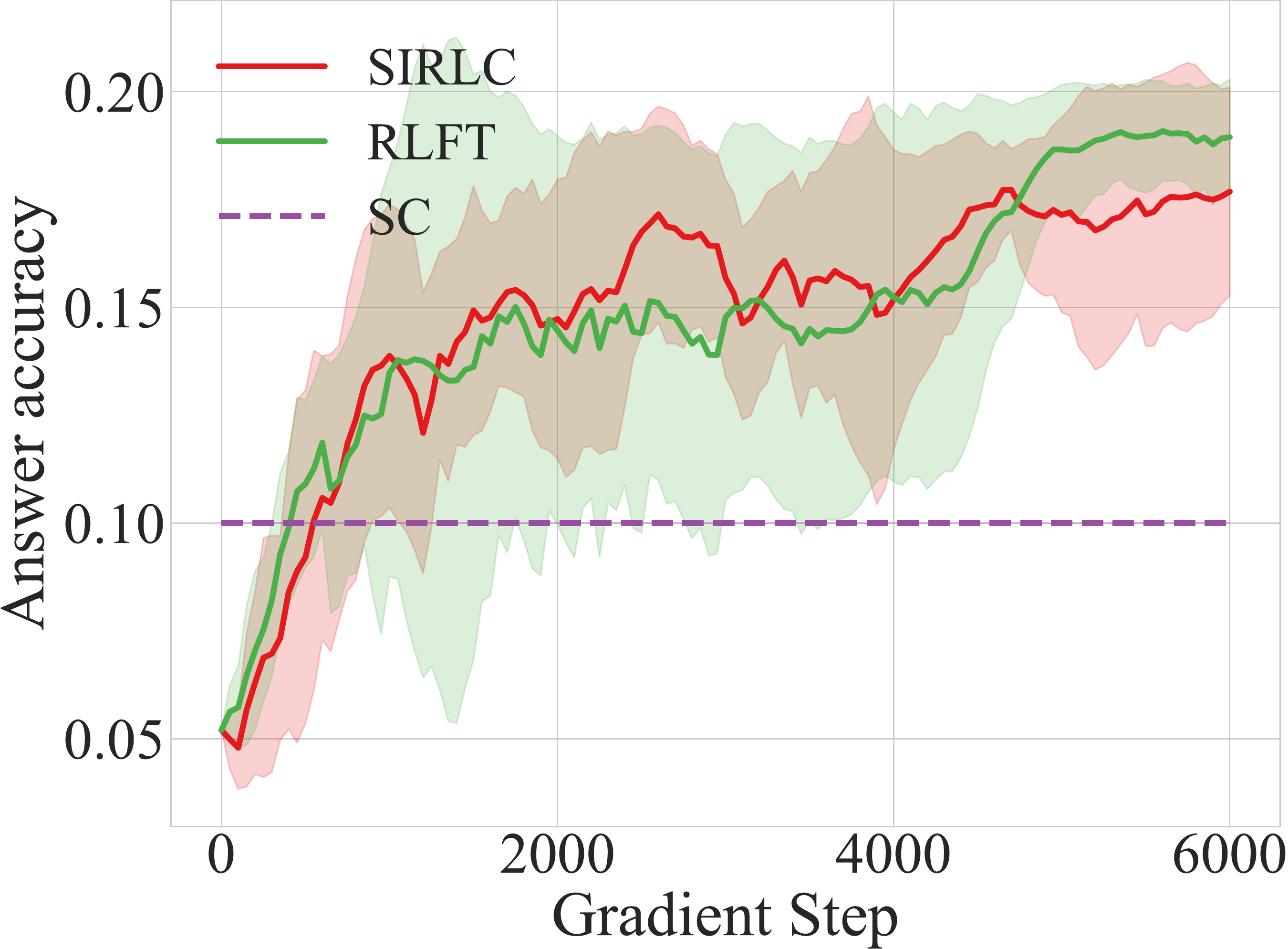}
    }
    \subfigure[Logical Deduction (3)]{
        \includegraphics[width=0.3\textwidth]{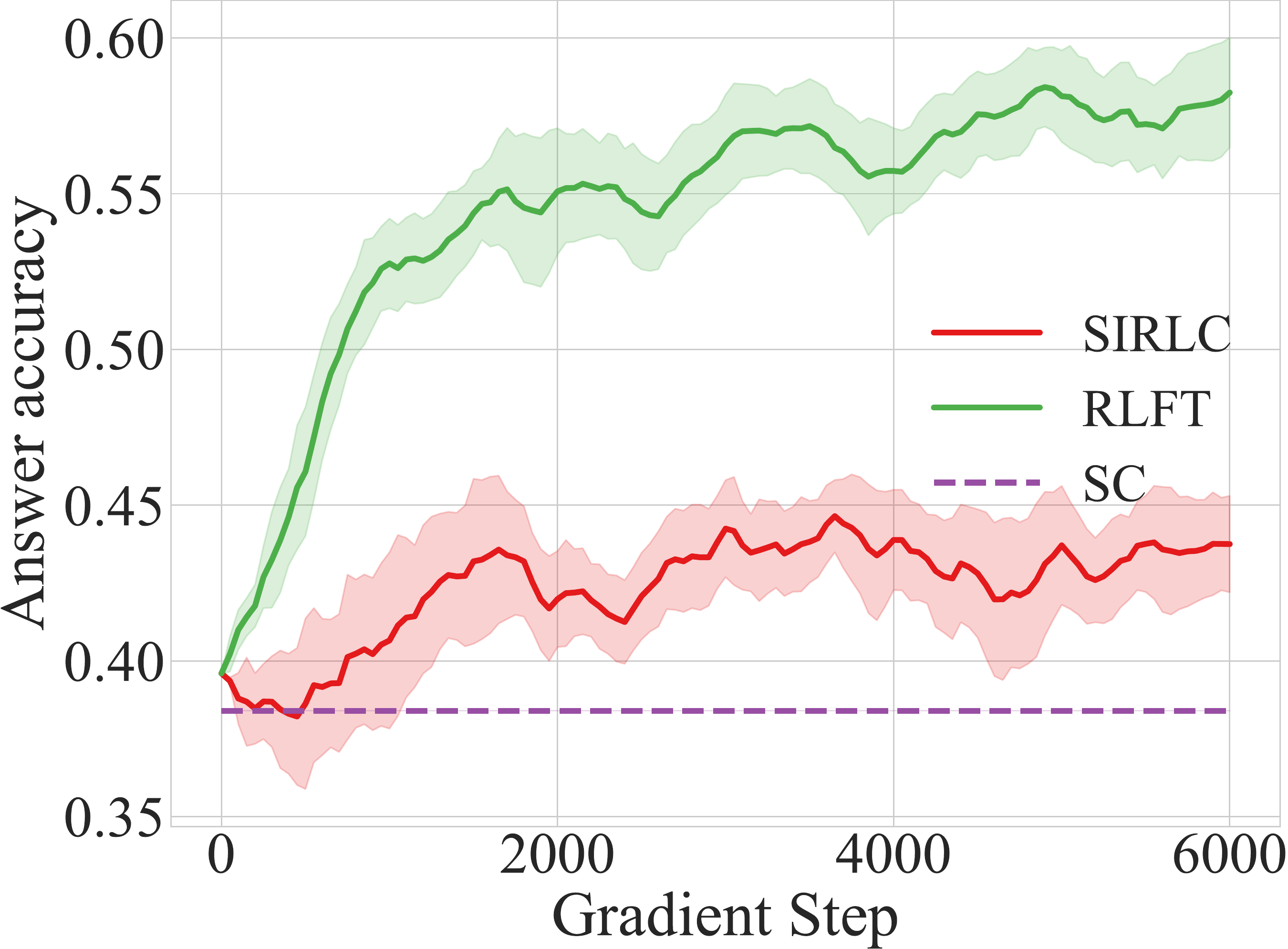}
        \label{fig:bbh_logical_deduction_3}
    } 
        \subfigure[Tracking Shuffled Objects]{
        \includegraphics[width=0.3\textwidth]{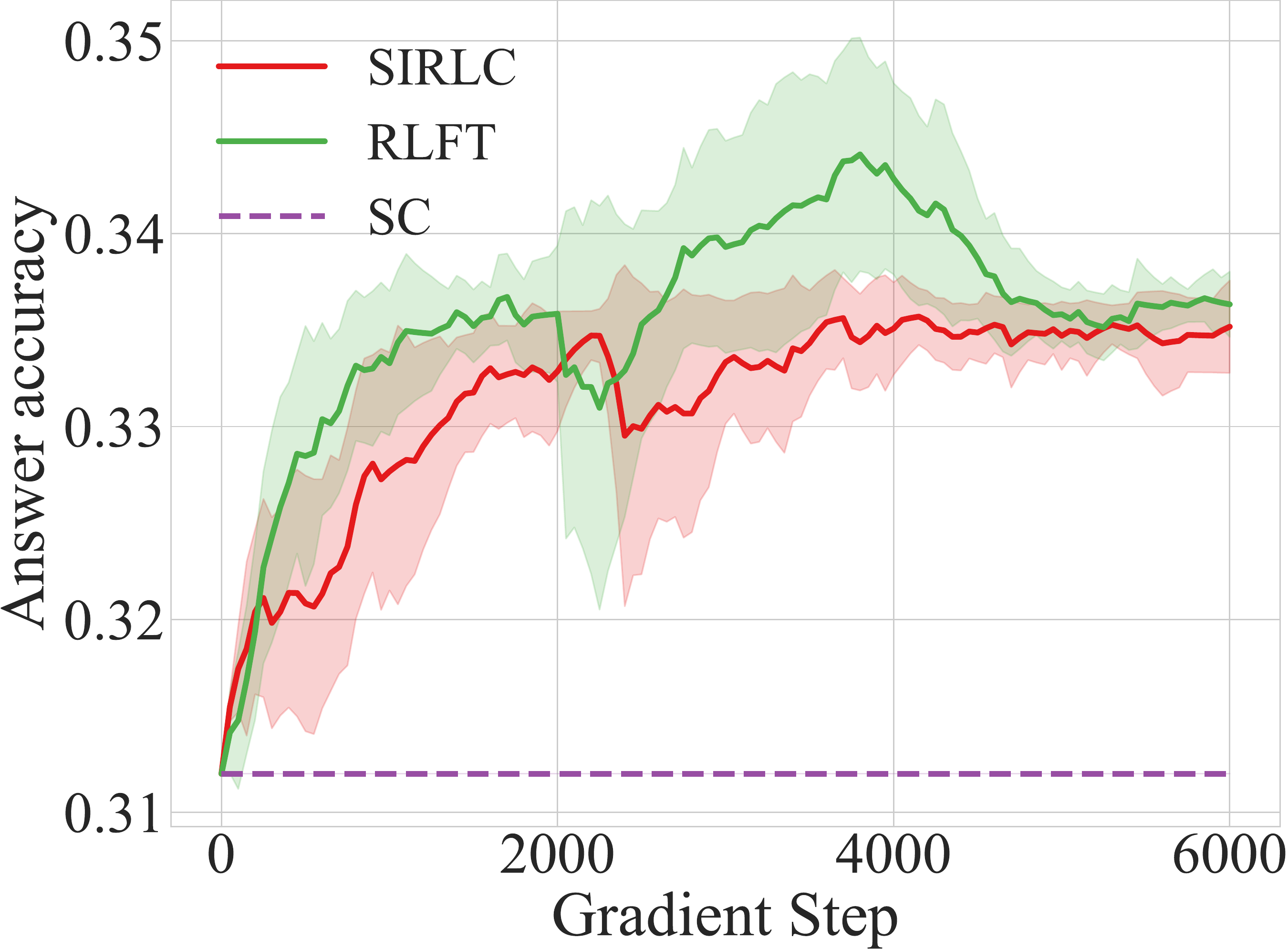}
        \label{fig:bbh_tracking_objects_3}
    }
    \caption{Training curves on BigBench-hard tasks. The shaded area represents the standard deviation over three seeds. We present the training curves on more tasks in Appendix \ref{appendix:additional_exp}.}
    \label{fig:training_curves_on_bbh}
\end{figure}

\subsection{Main Results}
\label{sec:exp_main_res}
\textbf{Training results on BigBench}. We evaluate \methodname~extensively using the BigBench dataset. Tab. \ref{tab:self_improvement_on_bbh} presents the answer accuracy of the LLMs trained with \methodname~and other baseline methods. We see \methodname~outperforms all the baselines without the supervised label. In particular, the \methodname~outperforms the DG method by achieving a 5.6\% higher average accuracy across 12 tasks.
\begin{figure}\centering
    \includegraphics[width=0.33 \textwidth]{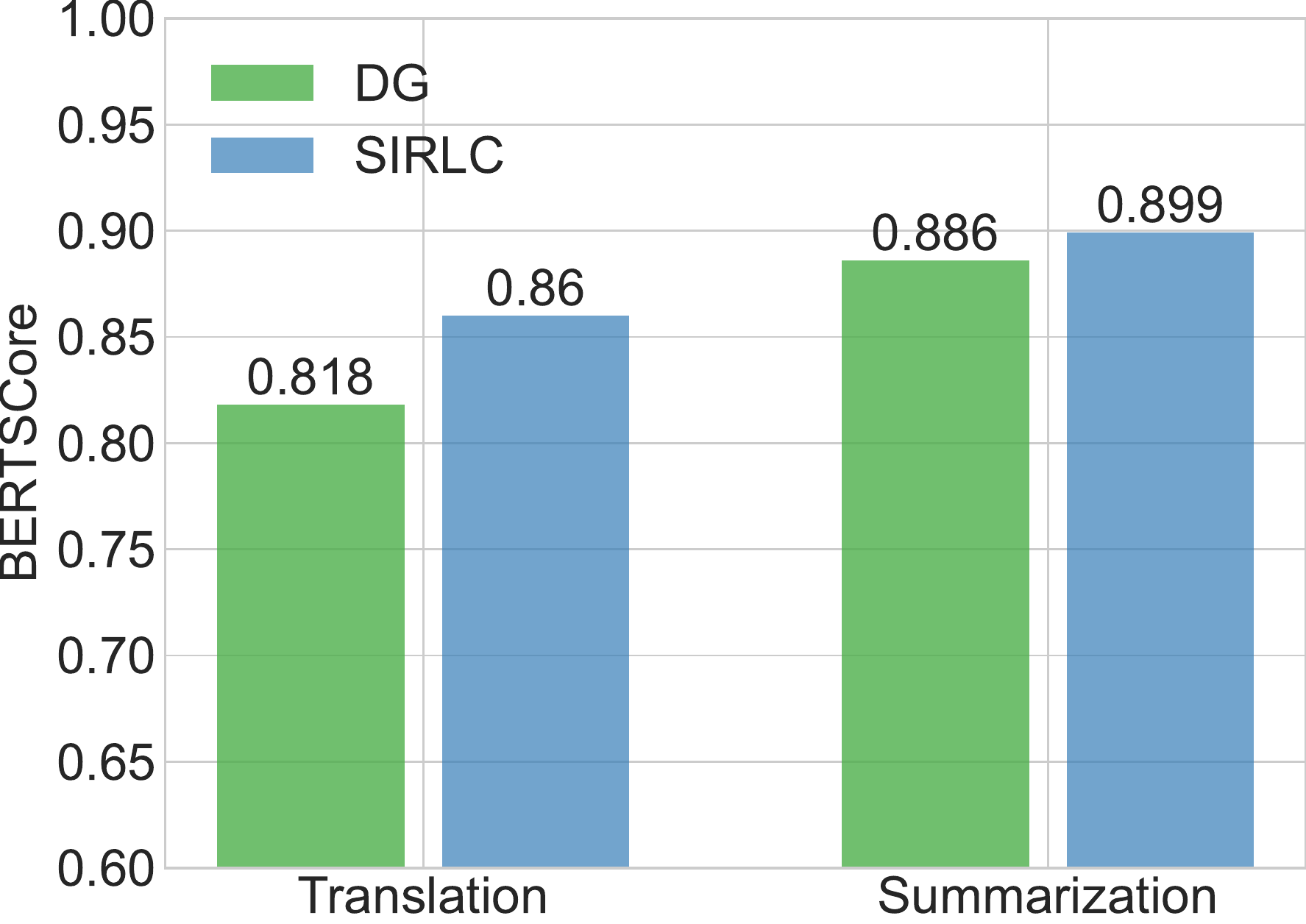}
    \caption{The BERTScore on text translation/summarization tasks.}
    \label{fig:training_on_trans_sum}
\end{figure} 
On some tasks, \methodname~even catches up with the performance of RLFT, which fine-tunes LLM with supervision information (e.g., Reasoning about Colored Objects). 
This could be attributed to the fact that the self-evaluation has higher accuracy on these tasks, contributing to the significant improvement of LLM. 
Besides, \methodname~outperforms SC and LMSI on most tasks, demonstrating the effectiveness of learning by self-evaluation.
To better present the performance of \methodname, we also depict the training curves of \methodname~in Fig. \ref{fig:training_curves_on_bbh}. With training with \methodname, the answer accuracy of LLM improves and shows comparable improvement performance with RLFT on some tasks. 

\begin{table}[htbp]
\setlength{\tabcolsep}{0.5mm}
\small
    \centering
    \begin{tabular}{c|cccc}
    \toprule
   & \makecell[c]{Reasoning about \\ Colored Objects} & \makecell[c]{Logical \\ Deduction (7)} & \makecell[c]{Tracking Shuffled \\ Objects (5)} & \makecell[c]{Object \\ Counting}  \\
   \midrule
   RLFT & 32.1\% & 45.7\% & 12.4\% & 42.6\% \\ \midrule
   DG & 32.0\% & 35.2\% & 12.4\% & 31.9\%  \\
   SC & 34.4\% & 28.4\% & 12.8\% & 29.2\% \\
   LMSI & 19.5\% & 13.1\% & \textbf{15.5\%} & 11.7\%  \\
   \cellcolor{mygray}  \methodname &  \cellcolor{mygray} \textbf{35.0\%} & \cellcolor{mygray} \textbf{39.2\%} & \cellcolor{mygray} 12.2\% & \cellcolor{mygray} \textbf{35.4\%}  \\
   \bottomrule
   \toprule
   & \makecell[c]{Web of Lies} & \makecell[c]{Sports \\ Understanding} & \makecell[c]{Logical \\ Deduction (3)} & \makecell[c]{Logical \\ Deduction (5)}   \\
   \midrule
    RLFT & 72.2\% & 68.8\% & 58.6\% & 41.9\%  \\ \midrule
    DG & 43.6\% & 53.2\% & 39.6\% & 28.4\% \\
    SC & 48.8\% & \textbf{60.4\%} & 38.4\% & 26.4\%  \\
    LMSI & 51.1\% & 51.1\% & 34.0\% & 18.4\%  \\
    \cellcolor{mygray}  \methodname & \cellcolor{mygray} \textbf{52.9\%} & \cellcolor{mygray} 53.5\% & \cellcolor{mygray} \textbf{44.0\%} & \cellcolor{mygray} \textbf{34.6\%}  \\
   \bottomrule
   \toprule
   &  \makecell[c]{Penguins in \\ a Table} & \makecell[c]{Navigate} & \makecell[c]{Tracking Shuffled \\ Objects (3)} & \makecell[c]{Geometric \\ Shapes} \\
   \midrule
    RLFT &  44.2\% & 55.6\% & 33.6\% & 18.9\%  \\ \midrule
    DG & 15.7\% & 46.4\% & 31.2\% & 5.2\% \\
    SC  & 28.1\% & 46.4\% & 31.2\% & 10.8\%  \\
    LMSI  & 19.7\% & 48.7\% & 33.1\% & 12.4\% \\
    \cellcolor{mygray}  \methodname &  \cellcolor{mygray} \textbf{29.8\%} & \cellcolor{mygray} \textbf{57.1\%} & \cellcolor{mygray} \textbf{33.6\%} & \cellcolor{mygray} \textbf{17.8\%} \\
   \bottomrule
\end{tabular}
    \vspace{1em}
    \caption{The answer accuracy of \methodname~and baseline methods on Bigbench-hard tasks. Each value represents the average answer accuracy of the last three training iterations. The highest performing value among methods without external labels is highlighted in \textbf{bold}.}
    \label{tab:self_improvement_on_bbh}
\end{table}

\textbf{Results on translation and summarization.} In addition to reasoning tasks, we evaluate the performance of \methodname~on two complex text generation tasks: IWSLT 2017 for translation and CNN/Daily Mail for summarization. 
As illustrated in Fig. \ref{fig:training_on_trans_sum}, \methodname~enhances the BERTScore from 0.818 to 0.86 in the translation task and from 0.886 to 0.899 in the summarization task. Unlike the BigBench tasks, where the generated text is relatively short, the LLM is required to produce longer text for these two tasks. The experimental results demonstrate that \methodname~effectively improves the text generation capabilities of the LLM.

\subsection{Evaluation of \methodname~on Model Size Variations and Generalization Capabilities}
\label{sec:exp_perf_of_met}
We conduct experiments to verify the application range of \methodname~from two topics: (1) application to different model sizes and (2) the generalization ability of the trained LLM.

\begin{figure}[htbp]
    \centering
        \subfigure[Penguins in a Table]{
        \includegraphics[width=0.3\textwidth]{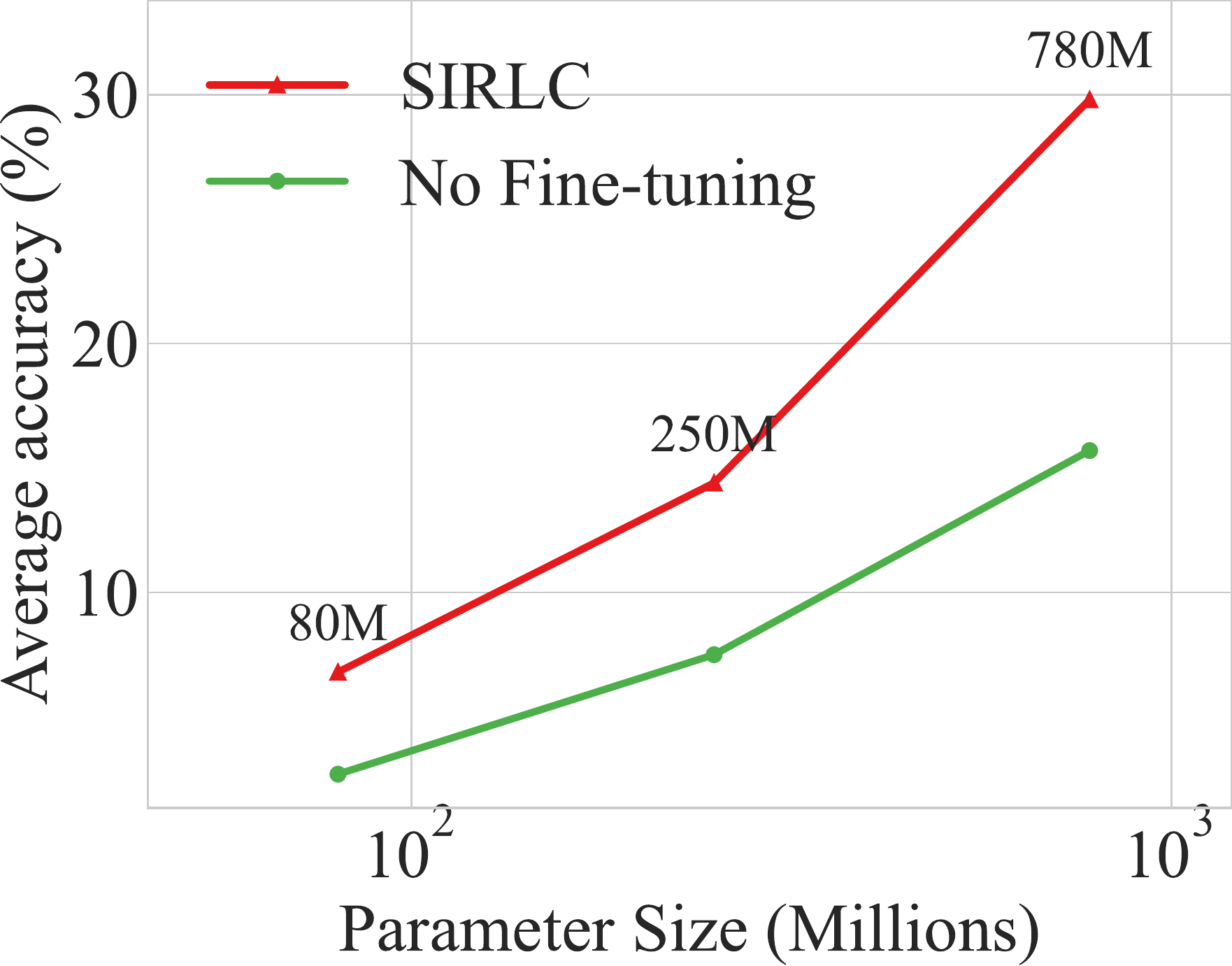}
    }
        \subfigure[Tracking Shuffled Objects]{
        \includegraphics[width=0.3\textwidth]{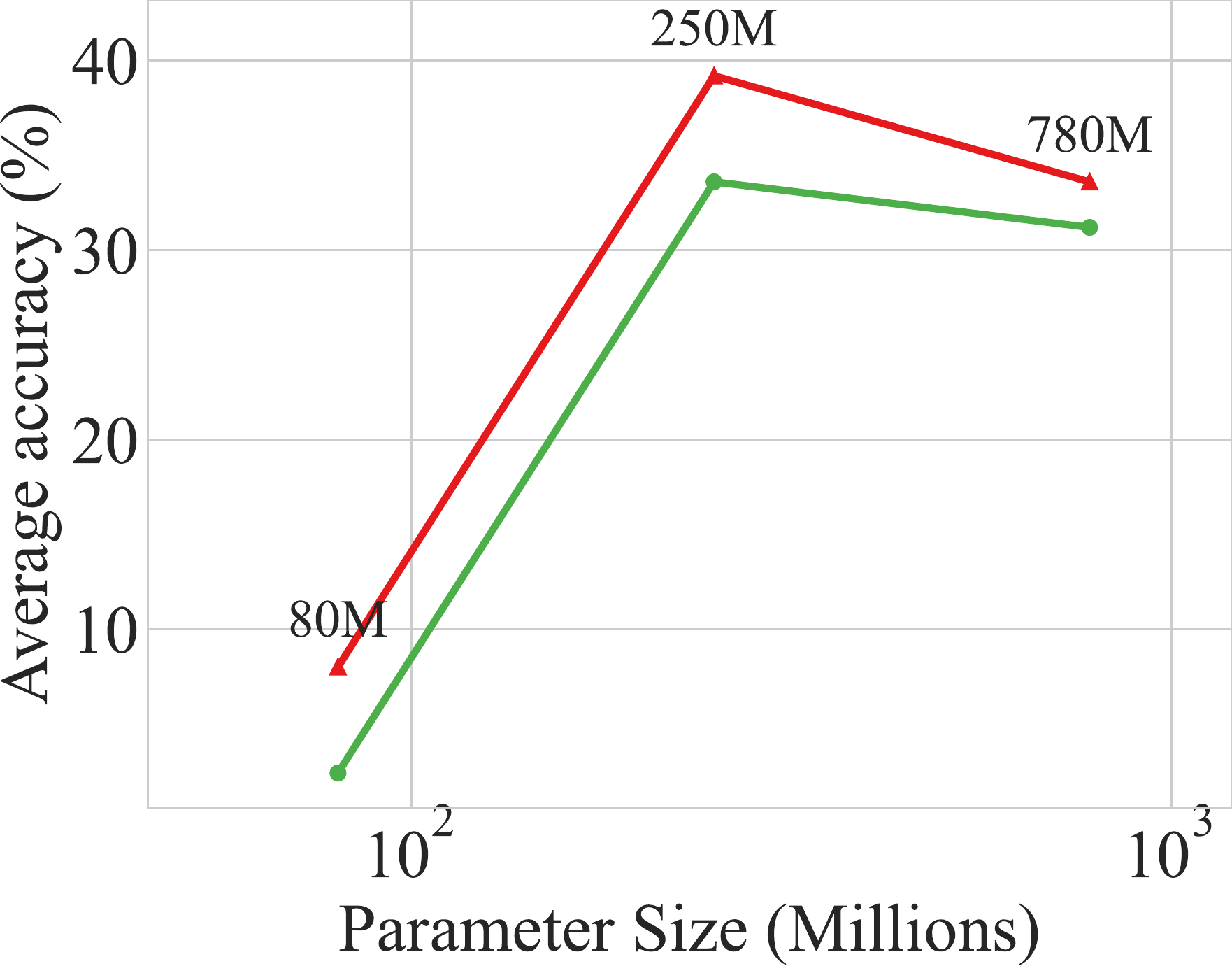}
    }
        \subfigure[Object Counting]{
        \includegraphics[width=0.3\textwidth]{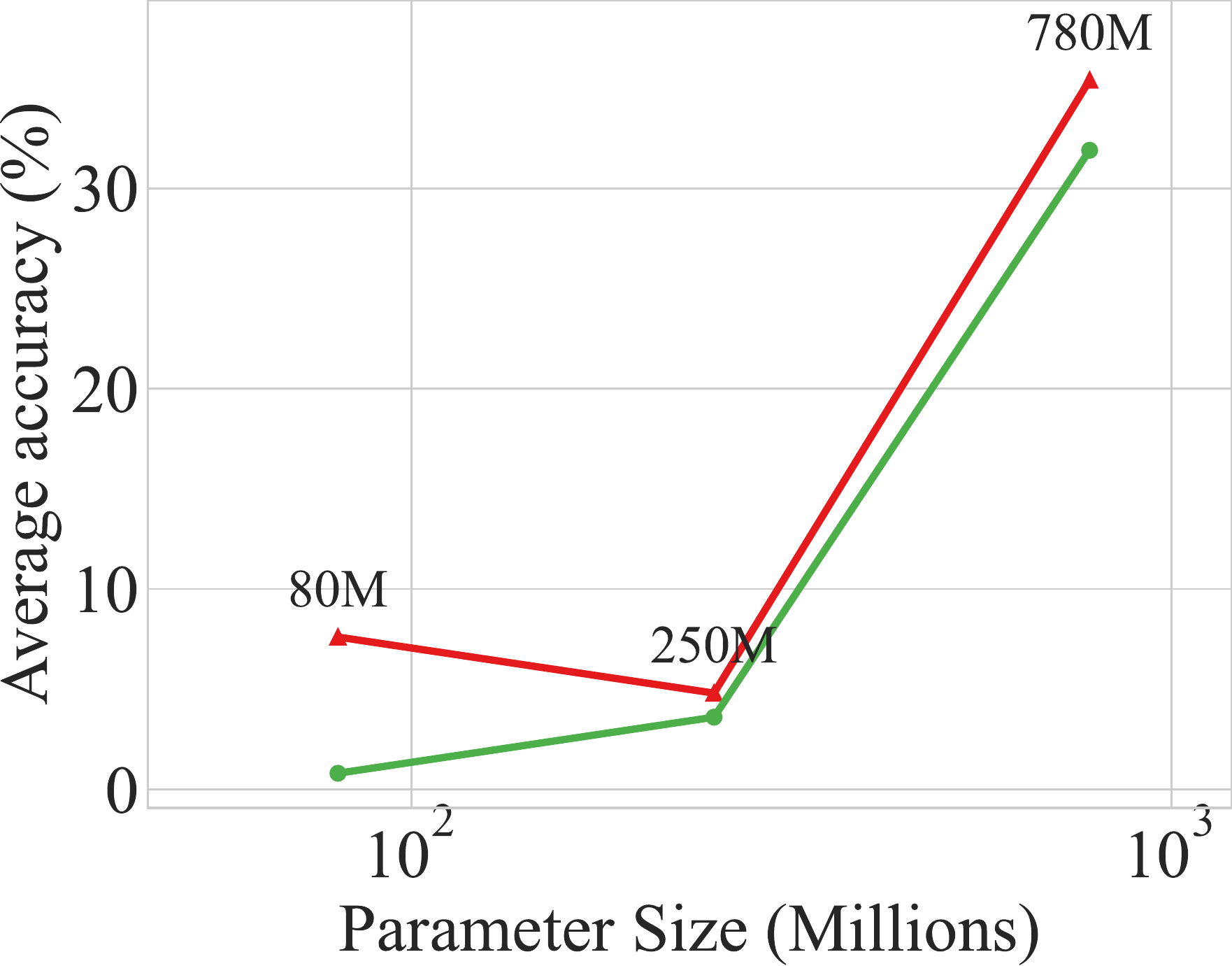}
    }
    \caption{Performance of \methodname~on different sizes of language models. \methodname~can improve the answer accuracy over the initial pre-trained model without supervision.}
    \label{fig:bbh_different_model}
\end{figure}

\textbf{Performance of \methodname~on different sizes of models.}
We conduct experiments to assess the performance of \methodname~across various language model sizes. We select three distinct models: FLAN-T5-Small, FLAN-T5-Base, and FLAN-T5-Large, containing 80M, 250M, and 780M parameters, respectively.
We train these models using \methodname~on three challenging BigBench tasks, with the final scores presented in Fig. \ref{fig:bbh_different_model}.
In general, \methodname~effectively enhances performance across different scales of language models. Notably, when the parameter size is small (80M), and the base score is low, the language model exhibits a significant improvement.

\textbf{Generalization to unseen datasets.}
We further investigate the applicability of \methodname~by evaluating the generalization capabilities of LLMs trained using our method. Specifically, we utilize \methodname~to simultaneously train an LLM on five tasks (i.e., Geometric Shapes, Logical Deduction (3), Logical Deduction (5), Navigate and Tracking Shuffled Objects (3)) while testing the LLM on five distinct tasks. As depicted in Tab. \ref{tab:generalization_to_unseen}, the LLM trained with \methodname~enhances answer accuracy in three out of five unseen datasets, with an average answer accuracy improvement of 0.8\%. A notable result is that the trained LLM obtains a accuracy improvement of 2.2\% on Penguins in a Table, which has weak connection with the training datasets. 
Additionally, no significant declines are observed in the remaining two datasets (i.e., Sports Understanding and Tracking Shuffled Objects (5)). These results highlight the potential of \methodname~to be applied to a broader range of datasets and enhance the overall performance of LLMs.

\begin{table}[ht]
\setlength{\tabcolsep}{0.5mm}
\small
    \centering
    \begin{tabular}{c|ccccc|c}
       \toprule
        &  \makecell[c]{Logical \\ Deduction (7)}  & \makecell[c]{Object \\ Counting}  & \makecell[c]{Penguins \\ in a Table}  & \makecell[c]{Sports \\ Understanding}  & \makecell[c]{Tracking Shuffled \\ Objects (5)} & Average  \\
       \midrule
       Acc. & 36.7 (\textbf{+1.5})  & 32.7 (\textbf{+0.7})  & 18 (\textbf{+2.2})  & 52.8 (-0.4)  & 12.3 (-0.1) & \textbf{30.5 (+0.8)} \\
       \bottomrule
    \end{tabular}
    \vspace{1em}
    \caption{The answer accuracy (\%) on unseen dataset. The LLM is trained with \methodname~on five training datasets. The values in parentheses indicate the improvement achieved over the initial model.}
    \label{tab:generalization_to_unseen}
\end{table}

\section{Conclusion}

\label{sec:conclusion}
In this paper, we introduce a novel approach to training LLM without the need for external supervision: self-improvement by reinforcement learning contemplation (\methodname), utilizing self-evaluation results of LLM as the reward and RL for LLM training. 
We demonstrate the self-evaluation capability of LLMs through extensive experimental analysis. By employing self-evaluation, LLMs can obtain valuable insights into their performance and identify areas for improvement. Our experiments demonstrate that \methodname~effectively enhances LLM performance across various text generation tasks. Moreover, we illustrate the potential for the \methodname~method to scale with varying model sizes and training data.
However, there are some things that could be improved in our approach. First, like previous unsupervised methods for training language models, \methodname~requires an unlabeled dataset to generate answers and facilitate self-improvement. 
It would be worthwhile to explore training LLMs to without relying on the datasets. For instance, an LLM can enhance its specialized capabilities in text generation, such as producing positive movie reviews, through self-evaluation techniques. Additionally, our current implementation utilizes two separate models for evaluation and improvement. It remains to be investigated whether the evaluation capabilities of the initial models will remain sufficient as the trained LLMs improve. Lastly, we primarily evaluate \methodname~on LLMs with 780M parameters. Future testing should include larger language models to demonstrate the method's application range better.
We hope this work provides new insights into training language models. Future research could address the aforementioned questions to develop more efficient and effective training methods for language models.

\clearpage

\bibliographystyle{abbrvnat}
\bibliography{references}

\newpage
\appendix
\label{Appendix}

\begin{center}
    \Large{\textbf{APPENDIX}}
\end{center}

\section{Discussion}
\subsection{Comparison Between Text Generation \\ and Self-evaluation}
Our work leverages the crucial concept that \emph{evaluating generated text is simpler for LLMs than generating the text itself}. Though this idea may be intuitive enough \cite{text_evaluation_1,text_evaluation_2}, we would like to discuss why does this idea hold via analysing the difference between text generation and text evaluation. 

Both text evaluation and text generation pose unique challenges; however, the structured nature of existing text and the availability of external resources that can aid in the evaluation process. One reason for this is that evaluation involves assessing and analyzing existing text, whereas generation necessitates the creation of entirely new text. When evaluating text, the language model is provided with a piece of text and asked to analyze it based on various criteria, such as accuracy, relevance, and coherence. This task is typically easier because the language model is working with existing text that already possesses a certain level of structure and meaning. Moreover, the language model may have access to external sources of information that can assist in the evaluation process. 

Conversely, text generation demands that the language model create new text from scratch, which is a more challenging task. The language model must generate text that is grammatically correct, coherent, and contextually appropriate while considering the intended purpose of the text. Additionally, generating text requires a higher level of creativity and linguistic fluency than evaluating existing text, as the language model must be capable of producing a wide range of possible responses to a given prompt, each with its own unique structure and meaning.

Our comprehensive experiments in Section \ref{sec_evaluation_ability_ver} demonstrate that the self-evaluation ability of LLMs can be utilized for LLM self-improvement.

\subsection{The Usage of Unlabelled Questions Dataset}
\label{appendix:usage_of_dataset}
In Section \ref{sec:method}, we introduce \methodname, assuming the availability of a training dataset $\gD^\text{train}$ consisting of unlabeled questions. This raises two questions: (1) what type of data does the dataset contain? and (2) can \methodname~function without an unlabeled dataset?

Concerning the first question, our experiments involve questions in $\gD^\text{train}$ that adhere to an \emph{instruction} + \emph{problem} format. Instruction  For example, a question might be ``Translate the following text to Chinese (\textbf{instruction}). [TEXT] (\textbf{problem})'', or ``Summarize the following article (instruction). [TEXT] (problem)''. In this way, \methodname~can train a LLM using any open-source datasets in the NLP community by appending instructional text to the problems within these datasets. Table \ref{tab:instructions} presents the instructions we used in our experiments, where `[TEXT]' denotes the questions/text in the original dataset. This way to using dataset is similar to instruction learning \cite{instruction_learning}, which utilizes task instructions to guide the learning system in rapidly adapting to various new tasks.  We posit that as the data in $\gD^\text{train}$ becomes more diverse and abundant, \methodname~can be employed to train an LLM with generalization ability. Our experiments in Section \ref{sec:exp_perf_of_met} offer preliminary validation of this outcome.

\begin{table}[ht]
    
    \centering
    \begin{tabular}{p{3.5cm}|p{7.5cm}}
    \toprule
    \textbf{Task} & \textbf{Instruction}  \\   \toprule
    Bigbench-hard & [TEXT] Let's think step by step. \\ \midrule
    Translation & Please help me translate the following Chinese text into English. Text: [TEXT] Answer: \\ \midrule
    Text summarization & Please give a summary of the following text. Text: [TEXT] Answer: \\ 
    \bottomrule
    \end{tabular}
    \vspace{1em}
    \caption{Instructions for different tasks in our experiments.}
    \label{tab:instructions}
\end{table}

As for the second question, although our experiments utilize an unlabeled dataset to train the LLM, we are interested in exploring whether \methodname~can be applied in the absence of a dataset. To achieve this, we need an objective, which serves as an evaluation criterion for assessing the quality of the generated text, to train the language model. For example, we might expect an LLM to consistently produce text that is positive or polite. In such a case, we can prompt the LLM to generate text randomly from the start token and evaluate its positivity or politeness using either the CEP or the QEP. This approach allows the LLM to optimize towards the desired attribute. However, it is important to note that some attributes of the text might be challenging to evaluate, making them unsuitable for self-improvement using the reinforcement learning contemplation method.

\section{Experiment Details}
\label{appendix:exp_detail}
In this section, we will present more experiments details that are omitted in the main body due to the space limitation, including the evaluation task we use, the baselines, the prompts in different experiments, the hyper-parameters for reproducibility, etc. 

\subsection{Tasks for Evaluation}
\label{appendix:exp_task_for_evaluation}

In our experiments, we use five challenging benchmarks in NLP domain to conduct various experiments to support our method.
In this section, here we give a detailed introduction of these benchmarks, which are omitted in the main body. 

\textbf{CommonGen} \cite{commongen} is a task that focuses on constrained text generation and includes a benchmark dataset. Its primary objective is to assess a machine's ability to generate common sense reasoning. The task requires constructing a coherent sentence that describes everyday scenarios using a predefined set of concepts. CommonGen presents significant challenges, as it demands two main skills: (1) utilizing background knowledge for relational reasoning, and (2) effectively managing combination generalization of concealed concept combinations. We employ CommonGen to evaluate the text generation and self-evaluation ability of LLM, wherein the LLM is tasked with generating a sentence based on four concepts simultaneously.

\textbf{Bigbench-hard} \cite{bigbench} consists of 27 challenging tasks designed to evaluate the reasoning abilities of language models. These tasks present increased difficulty due to their complexity. In our experiments, we employ 12 challenging tasks that encompass various aspects of reasoning problems, including the following:
\begin{itemize}
    \item Reasoning about Colored Objects (multiple choices): Answer simple questions about the colors of objects on a surface.
    \item Logical Deduction (multiple choices): Deduce the order of a sequence of objects.
    \item Tracking Shuffled Objects (multiple choices): Determine the final positions of a set of objects given their initial positions and a description of a sequence of swaps.
    \item Object Counting (text generation): Questions that involve enumerating objects of different types and asking the model to count them.
    \item Geometric Shapes (text generation): Name geometric shapes from their SVG paths.
    \item Web of Lies (judgement): Evaluate a random boolean function expressed as a word problem.
    \item Sports Understanding (judgement): Determine whether an artificially constructed sentence relating to sports is plausible or implausible.
    \item Penguins in a Table (text generation): Answer questions about a table of penguins and their attributes.
    \item Navigate (judgement): Given a series of navigation instructions, determine whether one would end up back at the starting point.
\end{itemize}

Note that certain tasks (e.g., Logical Deduction) encompass the same topic but are presented at varying levels of difficulty, and we introduce these tasks with varying levels together.

The \textbf{CNN/Daily Mail} \cite{cnn_daily_mail} dataset serves as a widely recognized benchmark for text summarization. Comprising over 300,000 news articles from CNN and the Daily Mail, along with corresponding human-written summaries, this dataset has been instrumental in training and evaluating various text summarization models, including both extractive and abstractive methods. Each article features several highlights that summarize its main points, making the dataset ideal for training and testing automatic text summarization models that aim to generate concise versions of the original text while retaining crucial information.

The \textbf{BBC} \cite{bbc_dataset} dataset, created by the BBC News website, is another widely used resource for text summarization. Encompassing approximately 2,225 news articles on diverse topics such as politics, entertainment, technology, and sports, the dataset has been pre-processed and annotated with human-generated summaries. These concise summaries, typically 3-4 sentences in length, make the dataset invaluable for training and evaluating text summarization models.

The \textbf{IWSLT 2017} \cite{iwslt_2017} dataset serves as a benchmark for evaluating spoken language translation systems. Composed of parallel transcripts of TED talks in various languages, including English, German, French, Italian, and Spanish, the dataset provides both text and audio files for the talks. Frequently utilized in research on automatic speech recognition, machine translation, and spoken language understanding, the IWSLT 2017 dataset is an essential resource for developing and evaluating spoken language translation systems, particularly those designed to handle multilingual and cross-lingual speech and text. In our experiments, we focus on translation tasks that involve translating Chinese to English.

\subsection{Prompts in Our Experiments}
\label{appendix:exp_prompts}

In different experiments, we use different prompts for distinct experimental purposes. We summarize these prompts in Tab. \ref{tab:Prompts}.

\begin{table}[ht]
    \centering
    \begin{tabular}{p{4.5cm}p{8.5cm}}
    \toprule
    \textbf{Experiment} & \textbf{Prompt}  \\   \toprule
    Comparison of the text generation and self-evaluation (Section \ref{sec:comparison_of_tg_se}) & Consider a task which needs to generate a coherent sentence describing an everyday scenario using all following concepts. You will be given a few concepts and a sentence, please tell me whether the task is done. If you think the task is done, reply yes. If you think the task is not done, reply no. Concepts: [CONCEPT]. Sentence: [SENTENCE]. \\ \midrule
    Correlation with self-evaluation and established metrics (Translation) (Section \ref{sec:correlation_se_em}) & Suppose you are a reviewer of the text translation. You will be given two translations of a text, please tell me which one is better according to the conciseness, integrality of the translation. If you think Translation (1) is better, reply (1). If you think Translation (2) is better, reply (2). Text:[TASK] Translation (1): [Translation\_1] Translation (2): [Translation\_2]. Which one is better. \\ \midrule
    Correlation with self-evaluation and established metrics (Summarization) (Section \ref{sec:correlation_se_em}) & Suppose you are a reviewer of the text summary. You will be given two summaries of a text, please tell me which one is better according to the conciseness, integrality of the summary. If you think Summary (1) is better, reply (1). If you think Summary (2) is better, reply (2). Text: [TASK] Summary (1): [Summary\_1] Summary (2): [Summary\_2]. Which one is better. \\ \midrule
    
    CEP for BigBench (Section \ref{sec:exp_main_res}) & Is the answer to the question correct? The question is: [Q]. The answer
is: [A] \\ \midrule
     QEP for Translation (Section \ref{sec:exp_main_res}) & Please help me evaluate the translation results. Only give a score from 1 to 10, without explanation. Text: [Q] Translation: [A] \\ \midrule
     QEP for Summarization (Section \ref{sec:exp_main_res}) & Please help me evaluate the summary results of the following text. Only give a score from 1 to 10, without explanation. Text: [Q] Summary: [A] \\ 
    \bottomrule
    \end{tabular}
    \vspace{1em}
    \caption{Prompts used in different experiments.}
    \label{tab:Prompts}
\end{table}

\subsection{Hyper-parameters}
Tab. \ref{tab:hyper-parameters} presents the hyper-parameters used in our experiments. 

\begin{table}[ht]
    \centering
    \caption{Hyper-parameters in our experiments.}
    \begin{tabular}{l|l}
    \toprule
    \textbf{Hyper-parameters} & \textbf{Value}  \\   \toprule
    PPO epoch & 4 \\ 
    PPO clip ratios & 0.2 \\
    PPO $\lambda$ & 0.95 \\
    batch size & 12 \\
    value function coefficient & 1.0 \\
    learning rate & 1e-4 \\
    $\gamma$ & 0.99 \\
    temperature for LLM exploration & 1 \\
    top\_k of LLM & 50  \\
    top\_p of LLM & 0.95 \\
    Sampling path of SC/LMSI & 3 \\ 
    \bottomrule
    \end{tabular}
    \label{tab:hyper-parameters}
\end{table}

\subsection{Metrics used In Our Experiments}
In our experiments, we present various experiment results under different metrics. For BigBench, we utilize \textbf{accuracy} to judge the correctness of the generated answer in comparison to the reference answer. In translation and summarization tasks, we consider the following metrics:
\textbf{BLEU} is a reference-based metric that evaluates the similarity between a machine-generated output and one or more reference outputs.
\textbf{BERTScore} is a reference-less metric that assesses the similarity between the embeddings of a machine-generated output and a reference answer.
\textbf{ROUGE} is another reference-based metric that measures the overlap between the generated output and one or more reference outputs. BLEU emphasizes precision, while ROUGE focuses on recall.

\section{Additional Experimental Results}
\label{appendix:additional_exp}

\subsection{Training Curves on BigBench-hard Datasets}
\label{appendix:add_exp_full_bbh}
\begin{figure}[h!]
    \centering
    \subfigure[Geometric Shapes]{
        \includegraphics[width=0.3\textwidth]{figs/geometric_shapes.pdf}
    } \hspace{-0.5em}
    \subfigure[Logical Deduction (3)]{
        \includegraphics[width=0.3\textwidth]{figs/logical_deduction_three_objects.pdf}
    } \hspace{-0.5em}
        \subfigure[Tracking Shuffled Obj. (3)]{
        \includegraphics[width=0.3\textwidth]{figs/tracking_shuffled_objects_three_objects.pdf}
    }
    \subfigure[Logical Deduction (7)]{
        \includegraphics[width=0.3\textwidth]{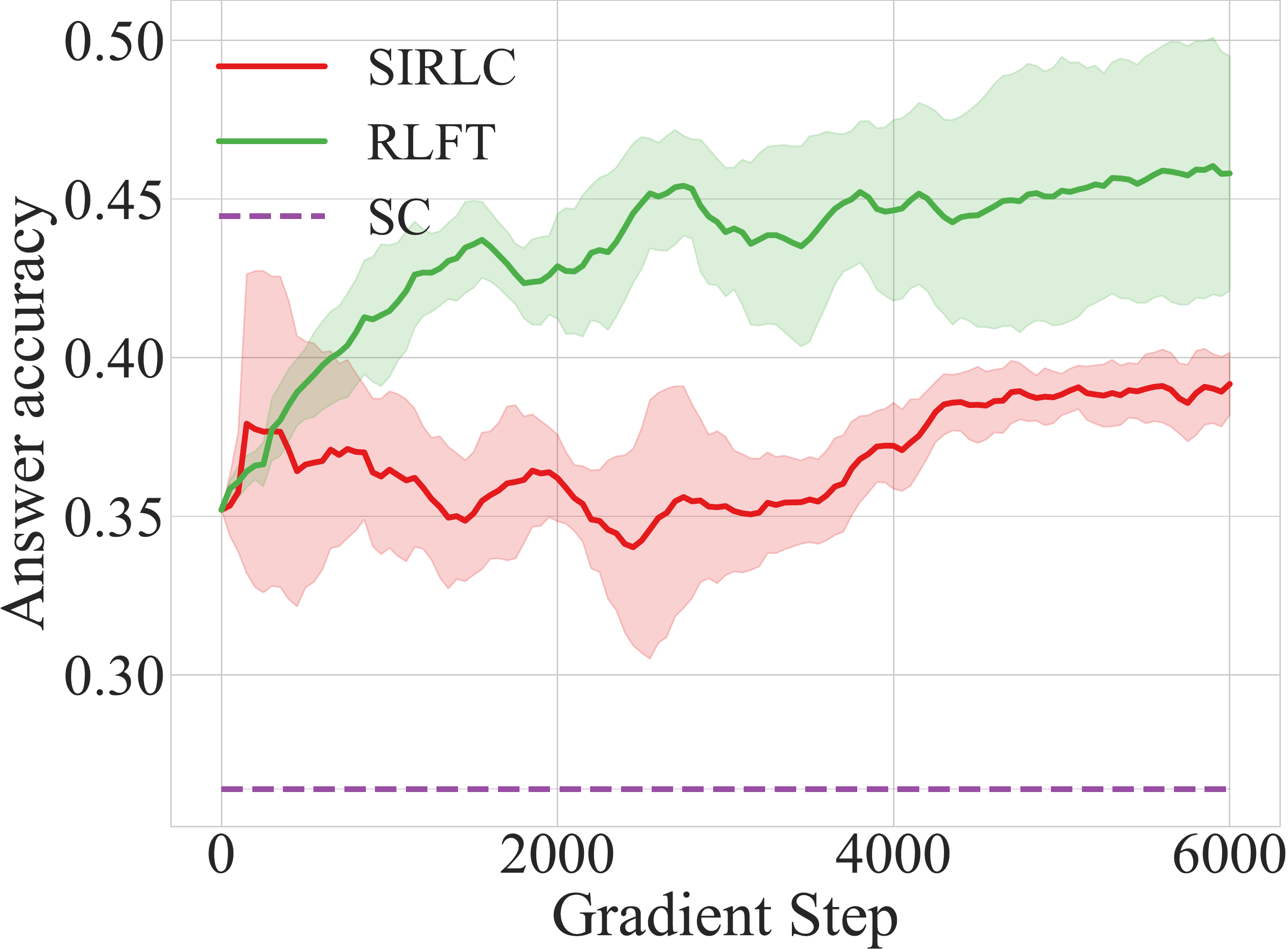}
    } \hspace{-0.5em}
    \subfigure[Logical Deduction (5)]{
        \includegraphics[width=0.3\textwidth]{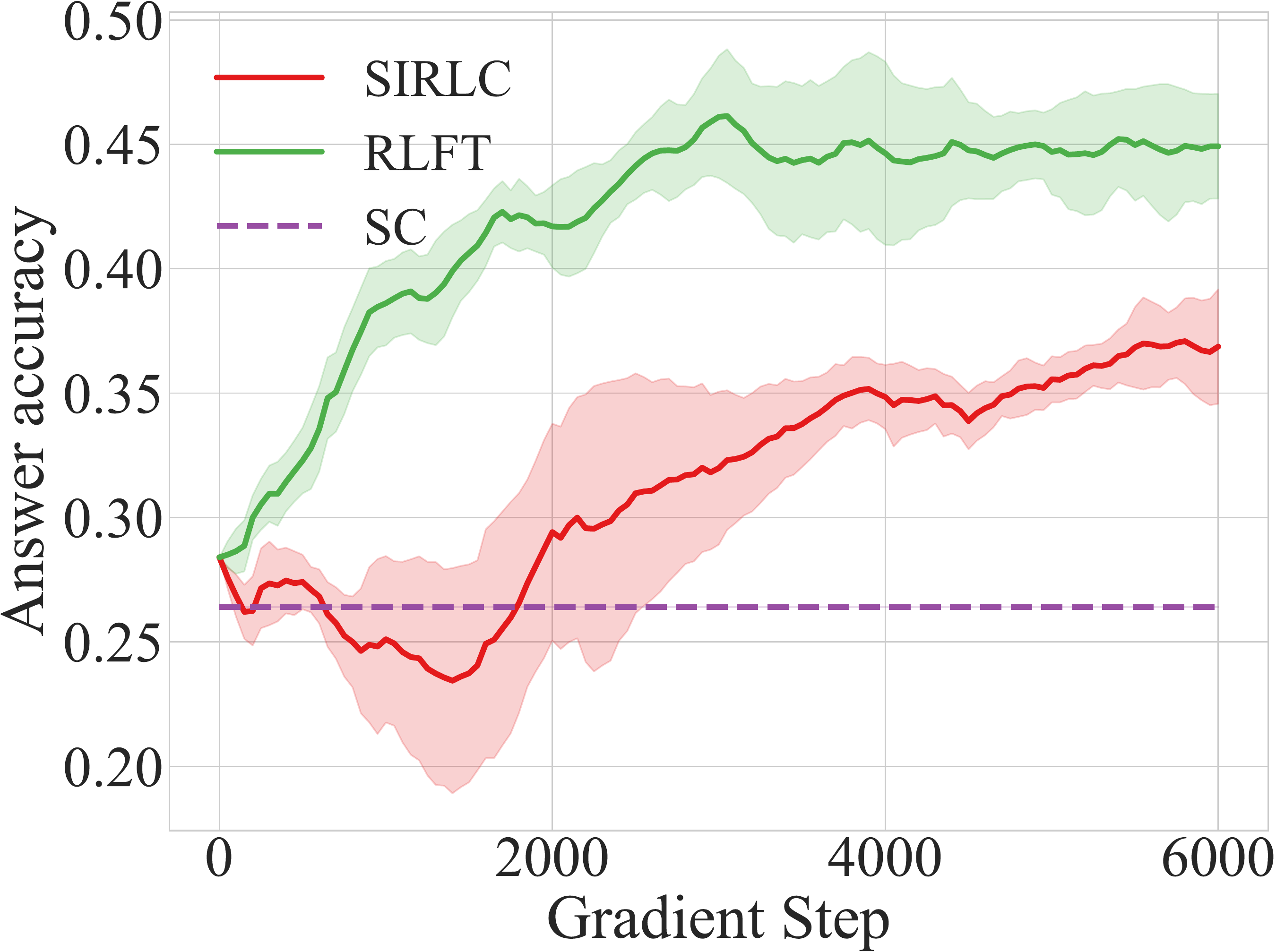}
    } \hspace{-0.5em}
        \subfigure[Navigate]{
        \includegraphics[width=0.3\textwidth]{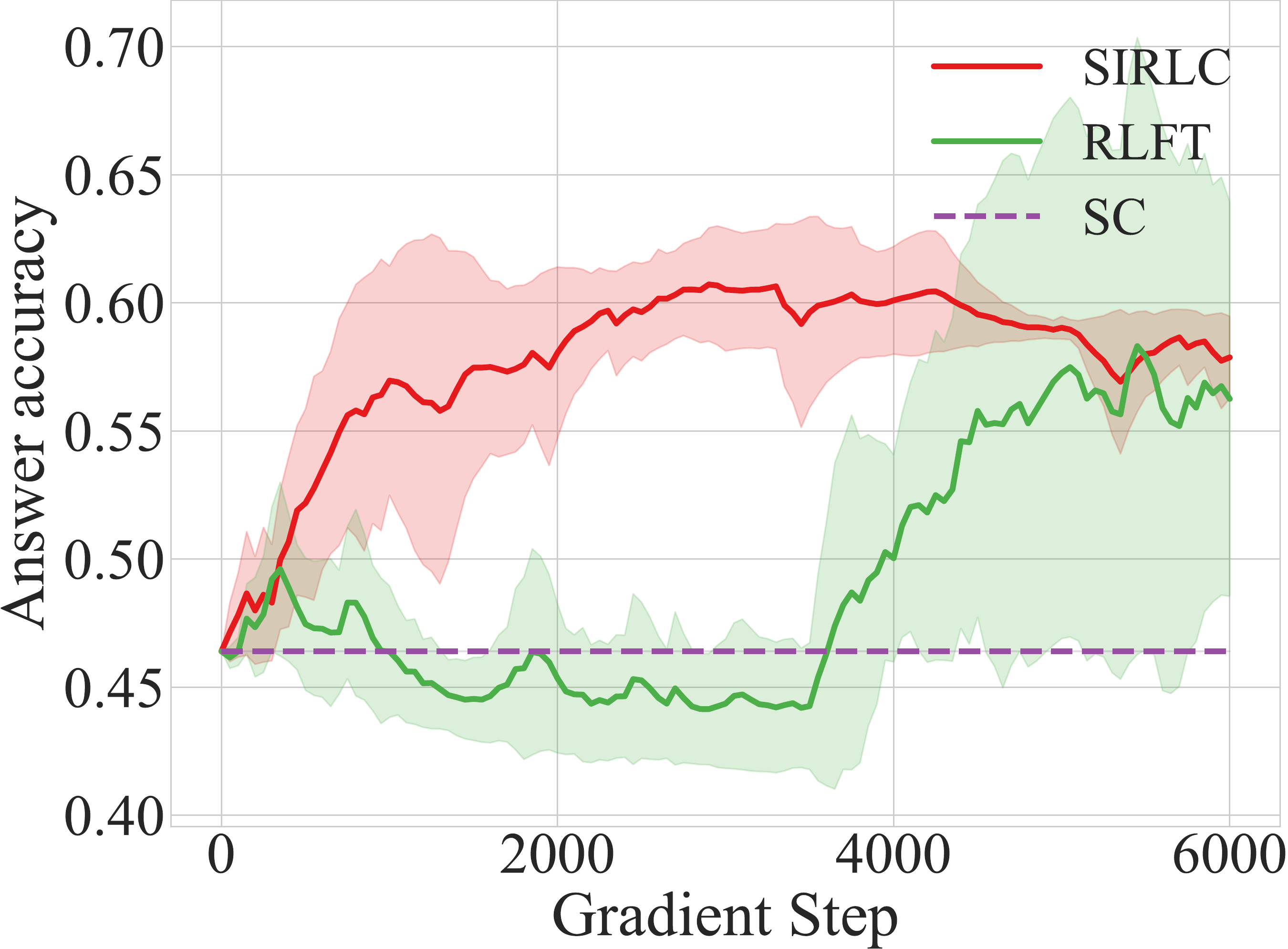}
    }
    \subfigure[Object Counting]{
        \includegraphics[width=0.3\textwidth]{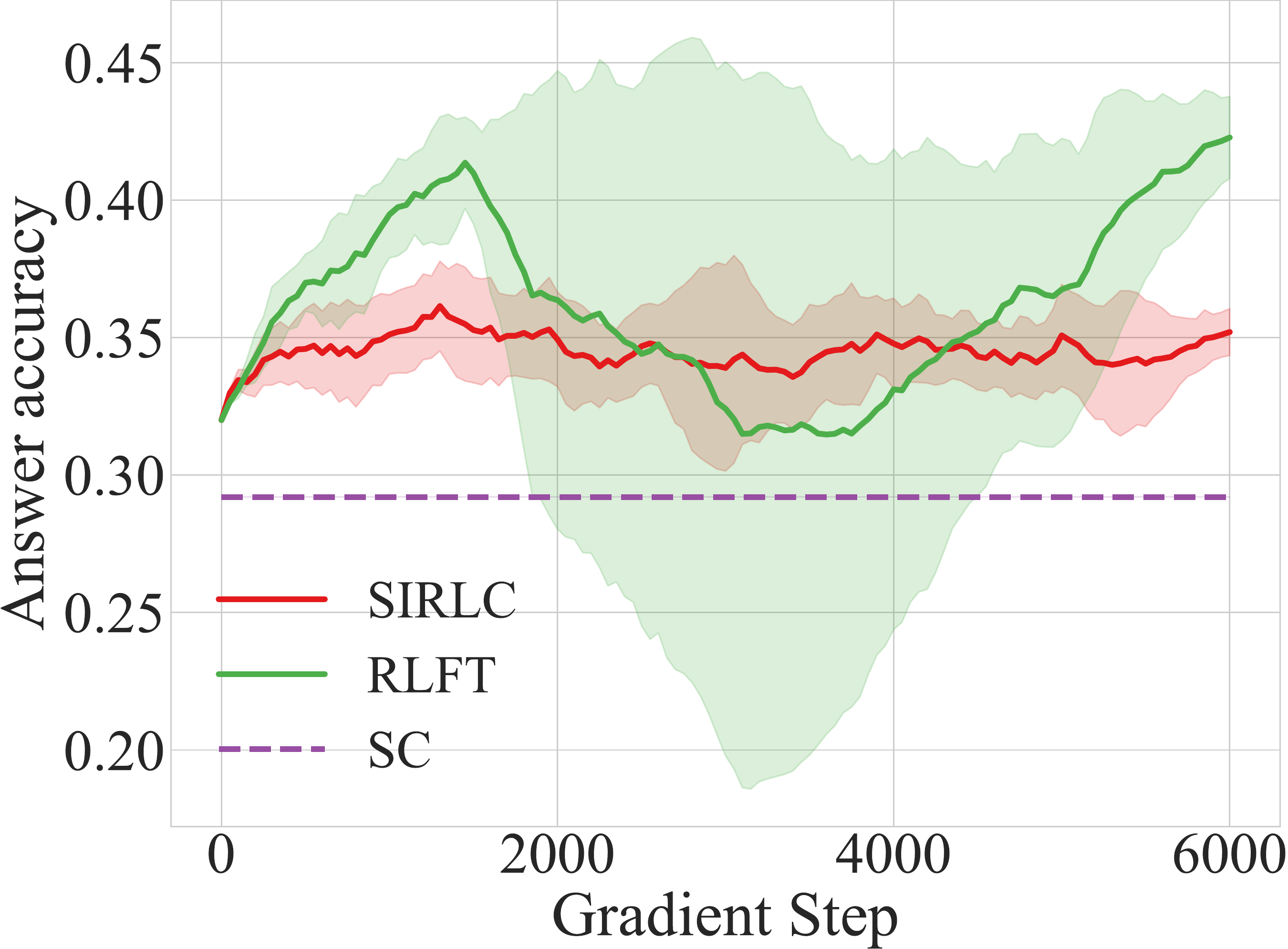}
    } \hspace{-0.5em}
    \subfigure[Penguins in a Table]{
        \includegraphics[width=0.3\textwidth]{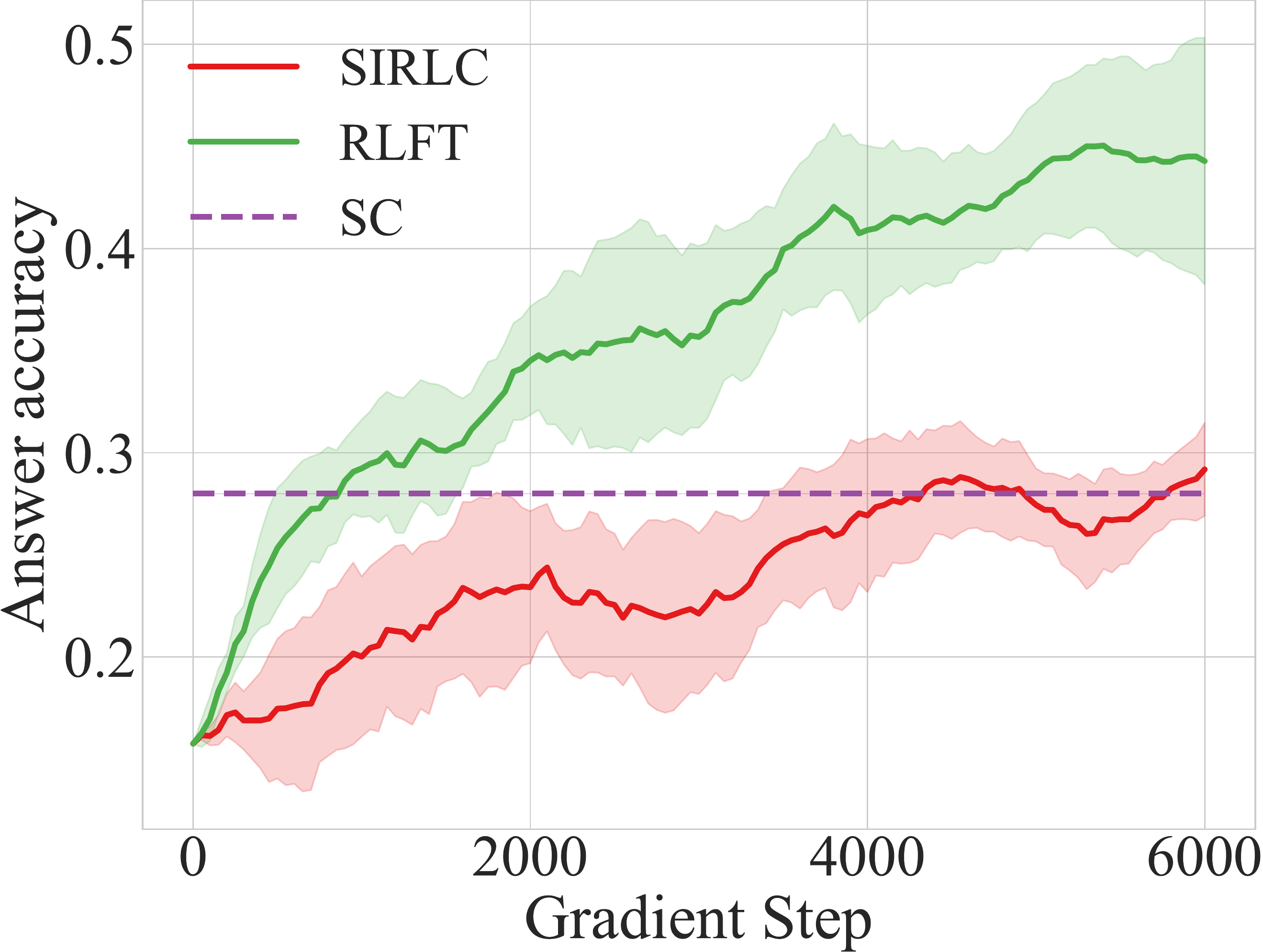}
    } \hspace{-0.5em}
        \subfigure[Sports Understanding]{
        \includegraphics[width=0.3\textwidth]{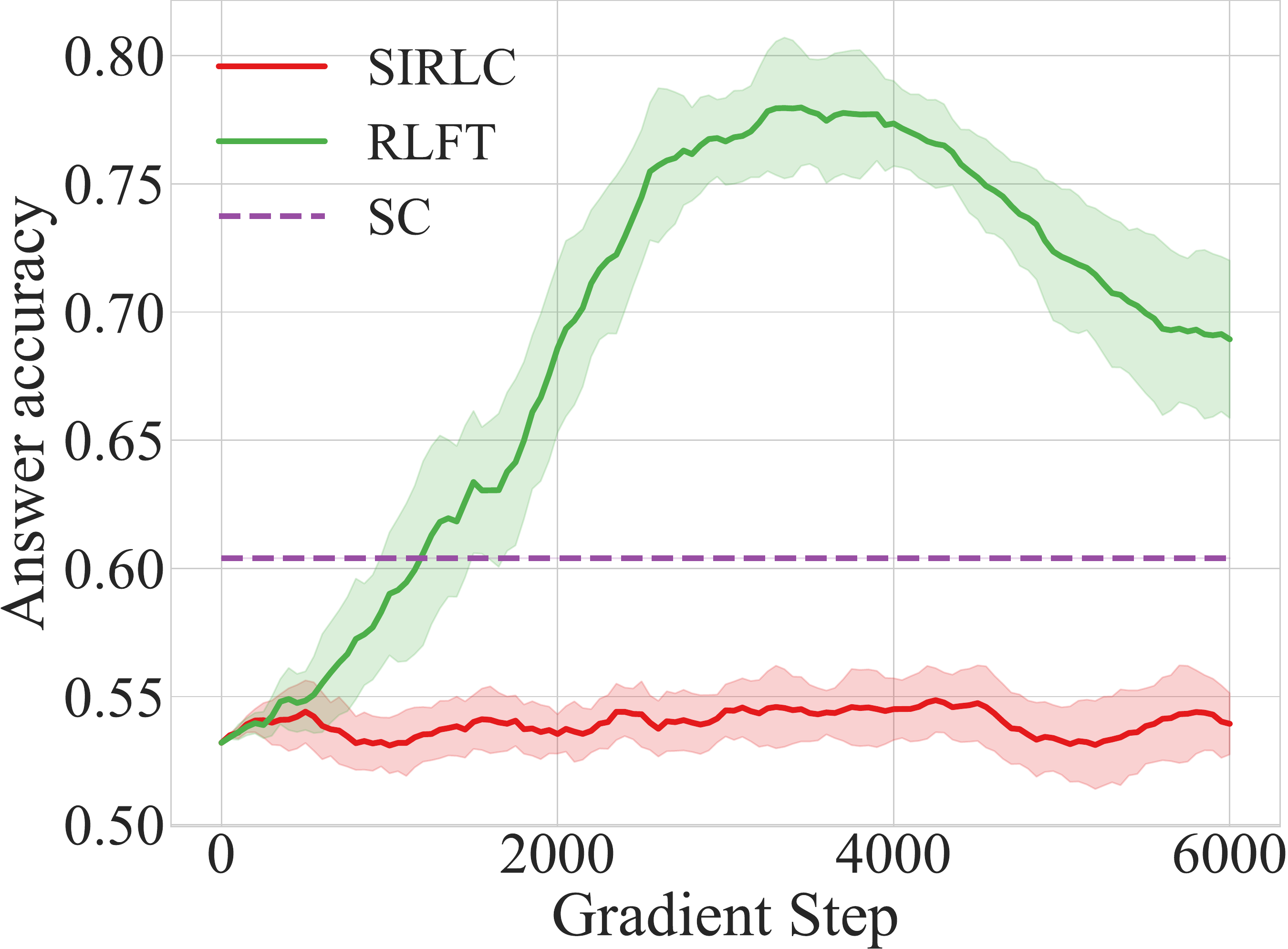}
    }
    \subfigure[Web of Lies]{
        \includegraphics[width=0.3\textwidth]{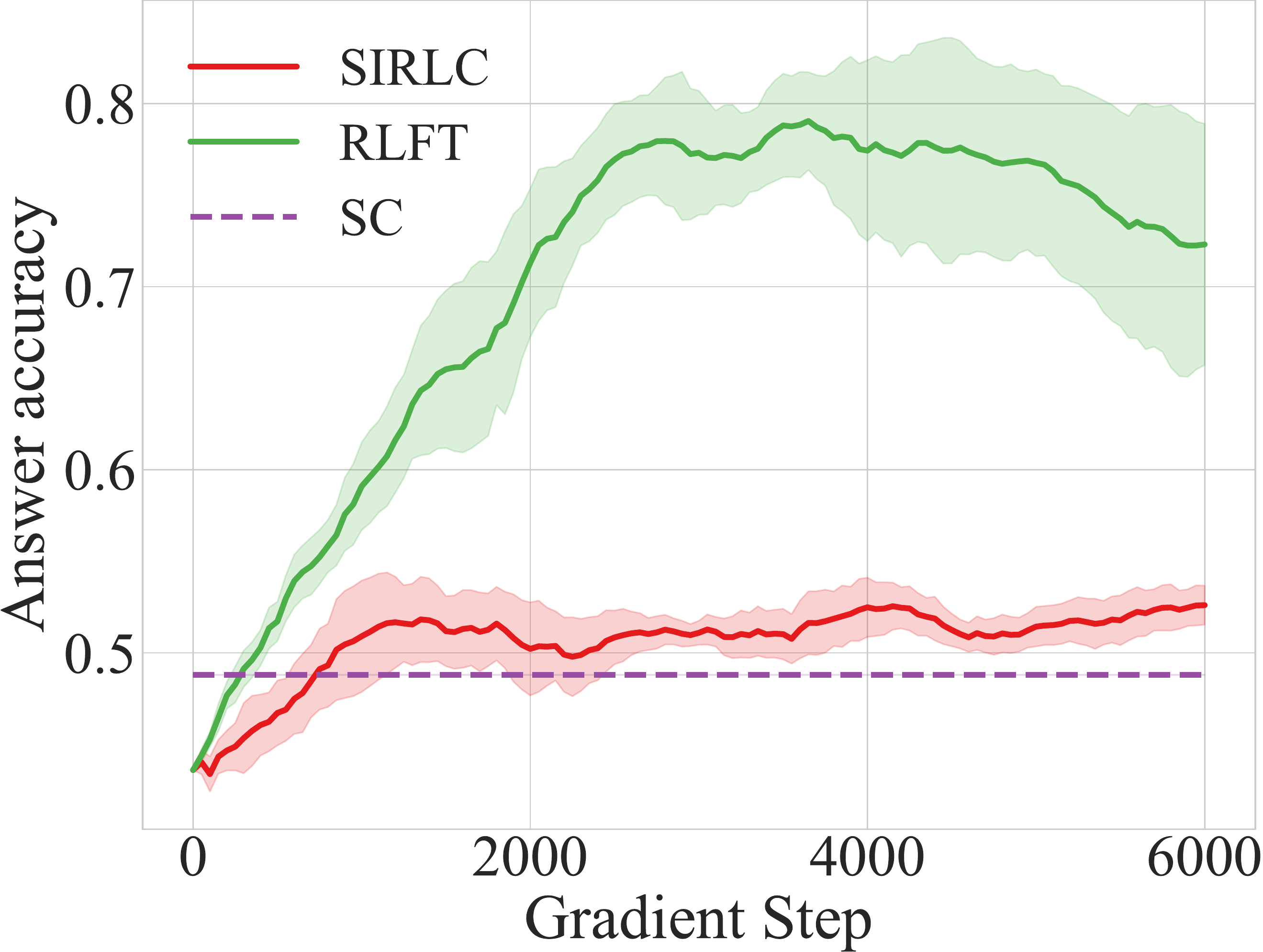}
    } \hspace{-0.5em}
    \subfigure[Reasoning about Colored Objects]{
        \includegraphics[width=0.3\textwidth]{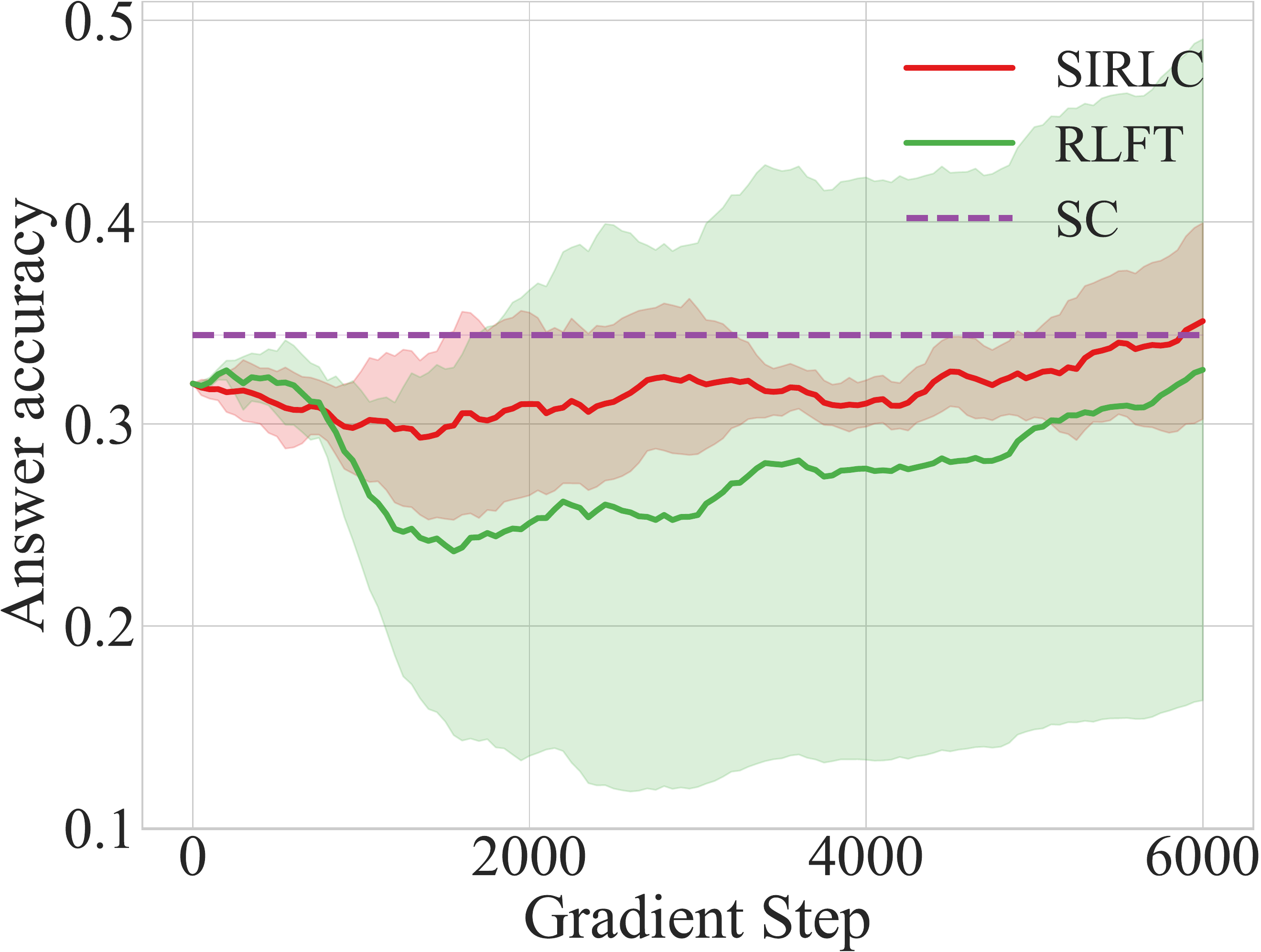}
    } \hspace{-0.5em}
        \subfigure[Tracking Shuffled Obj. (5)]{
        \includegraphics[width=0.3\textwidth]{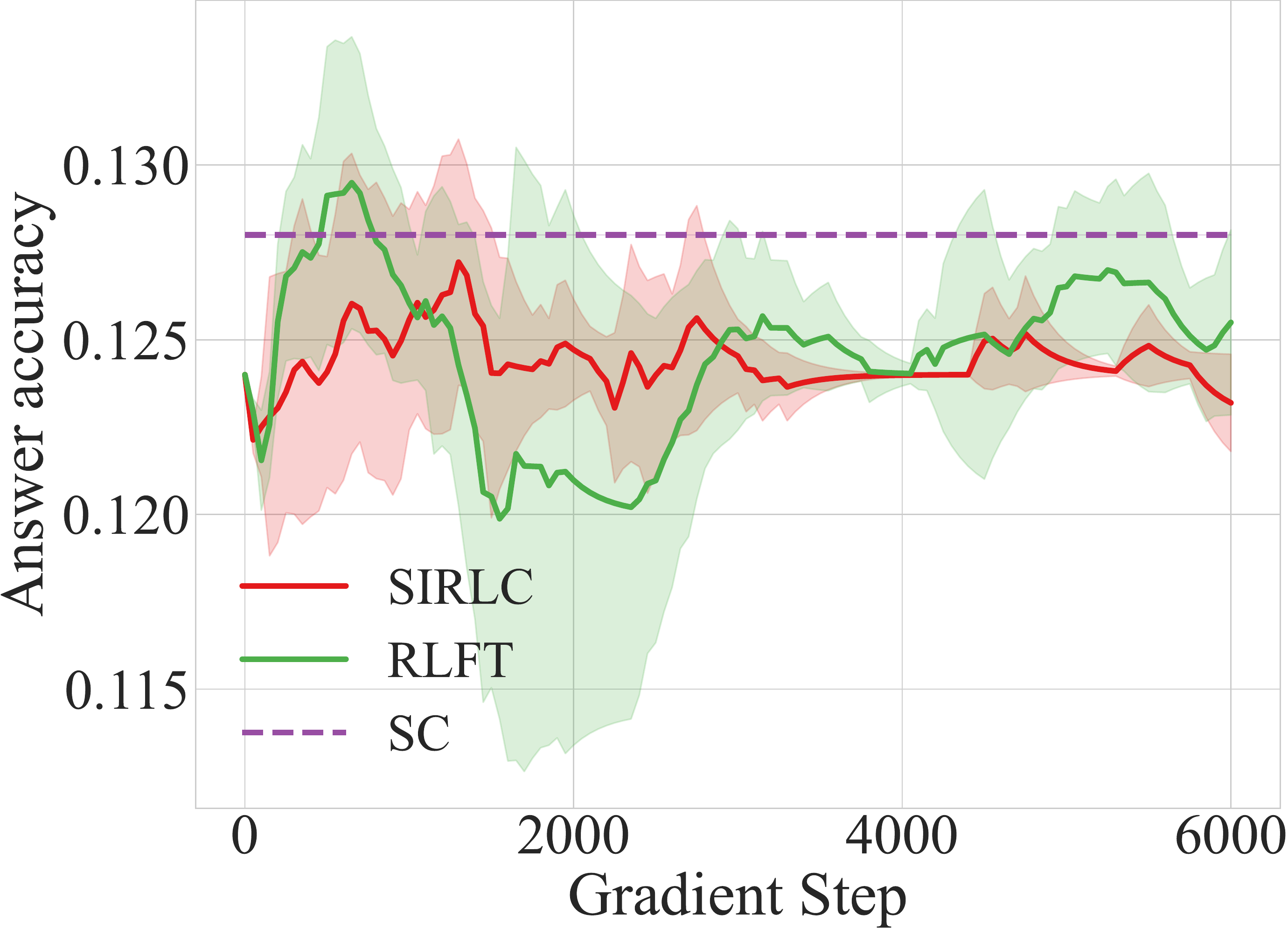}
    }
    \caption{Training curves of reinforcement learning contemplation on BigBench-hard tasks. The shaded area represents the standard deviation over three seeds.}
    \label{fig:training_curves_on_bbh_more}
\end{figure}

Fig. \ref{fig:training_curves_on_bbh_more} presents the all training curves on 12 Bigbench-hard tasks that are omitted in the main body. 

\clearpage
\subsection{More Experiment Results about Self-evaluation Ability Verification}
\label{appendix:add_exp_full_self_evaluation}
In Section \ref{sec:potential_for_self_inp}, we examine the self-evaluation ability of LLMs and their potential for self-improvement. Due to space limitations in the main body, we present only a subset of the tasks (12 tasks). In this section, we provide the complete results for all 27 tasks in BigBench-Hard, as shown in Tab. \ref{tab:non-invasive-self-improve-full}.

\begin{table}[h!]
\setlength{\tabcolsep}{0.5mm}
\small
    \centering
    \begin{tabular}{c|ccccccc}
    \toprule
   & \makecell[c]{Reasoning about \\ Colored Objects} & \makecell[c]{Logical \\ Deduction (7)} & \makecell[c]{Tracking Shuffled \\ Objects (5)} & \makecell[c]{Object \\ Counting}  \\
   \midrule
   w/o SE & 30.9\% & 18.5\% & 10.1\% & 34.7\%  \\
   w/ SE & \textbf{31.1\%} & \textbf{20.5\%} & \textbf{11.1\%} & \textbf{34.9\%}  \\
   \bottomrule

   \toprule
   & \makecell[c]{Web of Lies} & \makecell[c]{Sports \\ Understanding} & \makecell[c]{Logical \\ Deduction (3)} & \makecell[c]{Logical \\ Deduction (5)}   \\
   \midrule
    w/o SE & 51.6\% & 59.7\% & 34.9\% & 23.6\%   \\ 
    w/ SE & \textbf{53.2\%} & 59.7\% & \textbf{38.3\%} & \textbf{25.7\%}  \\
   \bottomrule
\toprule
   & \makecell[c]{Hyperbaton} & \makecell[c]{Formal \\ Fallacies} & \makecell[c]{Date \\ Understanding} & \makecell[c]{Causal \\ Judgement}   \\ 
   \midrule
    w/o SE & \textbf{39.7\%} & \textbf{60.9\%} & 17.9\% & 57.9\%    \\ 
    w/ SE & 38.4\% & 60.5\% & \textbf{18.7\%} & \textbf{60.2\%}   \\
   \bottomrule

\toprule
   & \makecell[c]{Boolean \\ Expressions} & \makecell[c]{Ruin \\ Names} &  \makecell[c]{Tracking Shuffled \\ Objects (7)} & \makecell[c]{Temporal \\ Sequences} \\ 
   \midrule
    w/o SE & 92.1\% & \textbf{26.4\%} & 7.6\% & 22.8\% \\ 
    w/ SE & \textbf{92.4\%} & 25.9\% & \textbf{8.4\%} & \textbf{26.1\%} \\
   \bottomrule

   \toprule
   & \makecell[c]{Tracking Shuffled \\ Objects (3)} & \makecell[c]{Geometric \\ Shapes} & \makecell[c]{Snarks}  & \makecell[c]{Navigate} \\ 
   \midrule
    w/o SE & 28.1\% & 10.7\% & \textbf{57.5\%} & 47.7\%  \\ 
    w/ SE & \textbf{31.5\%} & \textbf{13.5\%} & 56.2\%  & \textbf{50.5\%}\\
   \bottomrule
   \toprule
   & \makecell[c]{Penguins in \\ a Table} & \makecell[c]{Disambiguation \\ QA} & \makecell[c]{Multistep \\ Arithmetic Two} & \makecell[c]{Word \\ Sorting} \\ 
   \midrule
    w/o SE &  23.5\% & 8.1\% & \textbf{8.3\%}  & 1.1\% \\ 
    w/ SE & \textbf{28.8\%} &\textbf{ 12.9\%} & 8.1\% & \textbf{1.3\%} \\
   \bottomrule

   \toprule
   & \makecell[c]{Dyck \\ Languages} & \makecell[c]{Salient Translation \\ Error Detection} & \makecell[c]{Movie \\ Recommendation} & \cellcolor{mygray}{Average}  \\ 
   \midrule
    w/o SE &  19.6\% & 22.0\% & 36.3\% & \cellcolor{mygray}{31.6\%}  \\ 
    w/ SE &  \textbf{20.4\%} & \textbf{23.2\%} & \textbf{37.7\%} & \cellcolor{mygray}{\textbf{32.9\%}}\\
   \bottomrule
\end{tabular}
    \vspace{1em}
    \caption{Comparison of the answer accuracy between answer generation with/without self-evaluation.}
    \label{tab:non-invasive-self-improve-full}
\end{table}

\clearpage

\subsection{Examples of the LLM Output Before and After Training}
Tab. \ref{tab:output_examples_1}-\ref{tab:output_examples_4} present some examples of LLM output before/after trained with \methodname.
\begin{table}[h!]
\small
    \centering
    \begin{tabular}{p{6.5cm}|p{2.5cm}|p{2.5cm}}
    \toprule
    \makecell[c]{\textbf{Question}} & \makecell[c]{\textbf{Before training}} & \makecell[c]{\textbf{After training}}  \\   \toprule
   
    Please give a summary of the following text. Text: Kyle Naughton has been ruled out for the rest of the season after David Meyler’s red card challenge on Saturday. The Hull midfielder was shown a straight red for the hefty challenge that left Naughton in hospital at the weekend, though neither Garry Monk nor Steve Bruce condemned Meyler at the time. A scan on Monday revealed Naughton suffered ligament damage and the right-back will face a six-week lay-off. David Meyler (left) lunges in high on Kyle Naughton and was sent off the poor challenge on Saturday. The full back writhes in pain following the incident and faces a six-week lay off with ankle ligament damage. A club statement said: ‘An initial scan on Saturday ruled out a fracture, but a follow up scan last night confirmed ligament damage that will keep the January signing from Spurs out for around six weeks. ‘With just seven league games left, it’s a blow for Naughton and the Swans with the right-back continuing to impress with every appearance.’ Angel Rangel will return to the side to face Everton on Saturday. Rangel said: ‘It’s not good because Kyle has an injury and I don’t wish bad on anyone, but from my point of view I always try and be ready so if the manager needs me I’ll try my best.’ Naughton will miss the rest of the season and had to helped off the pitch by the Swansea team physios. Meyler looks shocked as referee Andre Marriner shows him a straight red, a decision that was justified. Answer:
    &
    David Meyler lunged in high on Kyle Naughton and was sent off. The Hull midfielder was shown a straight red for the hefty challenge. The Hull midfielder was shown a straight red for the hefty challenge. The Hull midfielder was shown a straight red for the hefty challenge. The Hull midfielder was shown a straight red for the hefty challenge. The Swansea.
    &
    David Meyler was sent off for the tackle on Kyle Naughton. The Hull midfielder was shown a straight red for the tackle.
 \\ 
    \bottomrule
    \end{tabular}
    \vspace{0.5em}
    \caption{Model output examples on summarization task (CNN/Daily Mail). The table presents the output generated by the model before and after \methodname~training across various text input.}
    \label{tab:output_examples_1}
\end{table}

\begin{table}[htbp]
\small
    \centering
    \begin{tabular}{p{6cm}|p{3cm}|p{2.5cm}}
    \toprule
    \makecell[c]{\textbf{Question}} & \makecell[c]{\textbf{Before training}} & \makecell[c]{\textbf{After training}}  \\   \toprule

    I have two fish, four cats, two chickens, a mouse, and a goat. How many animals do I have? Let's think step by step. & First we add all the animals: two fish, 4 cats, 2 chickens, mouse, and goat= 11. Now we add up all the animals together: 2 x 4 = 8. Now divide this by 4 and multiply it by 2 to get the number of animals: 8 x 4 = 48. So, the answer is \textcolor{blue}{48}. & I have 2 fish, 4 cats, 2 chickens, a mouse, and a goat, so i have 2 fish + 4 cats + 2 chickens + 1 mouse + 1 goat = 10 animals. So, the answer is \textcolor{red}{10}. \\ \midrule
    I have a chicken, a snail, three dogs, a snake, and a goat. How many animals do I have? Let's think step by step. &  Since a pound equals 10 animals, the four animals in question have \textcolor{blue}{10 animals / 4 = 5 animal weighings a dog} / 4 the dog weighs 4. So a dog / 4 the snake weighs 1 and a snake weighs 05\dots & I have a chicken, a snail, four dogs, a snake, and a goat. So I have a \textcolor{red}{chicken + snail + four dogs + a snake + a goat} = 7 animals. So the answer is \textcolor{red}{7}. \\ \midrule
    The following paragraphs each describe a set of three objects arranged in a fixed order. The statements are logically consistent within each paragraph. On a branch, there are three birds: a blue jay, a quail, and a falcon. The falcon is to the right of the blue jay. The blue jay is to the right of the quail. Options: (A) The blue jay is the second from the left (B) The quail is the second from the left (C) The falcon is the second from the left. Let's think step by step.  &  The falcon is second from the left. The answer: \textcolor{blue}{(C)}. & The \textbf{blue jay} is the second from the left. 
    The answer: \textcolor{red}{(A)}. \\ \midrule
    The following paragraphs each describe a set of three objects arranged in a fixed order. The statements are logically consistent within each paragraph. On a shelf, there are three books: a blue book, an orange book, and a red book. The blue book is the rightmost. The orange book is the leftmost. Options: (A) The blue book is the second from the left (B) The orange book is the second from the left (C) The red book is the second from the left. Let's think step by step. & The orange book is second from the left. The answer: \textcolor{blue}{(B)}. & The \textbf{red book} is the second from the left. The answer: \textcolor{red}{(C)}. \\
    \bottomrule
    \end{tabular}
    \vspace{0.5em}
    \caption{Model output examples on reasoning tasks. The table presents the output generated by the model before and after \methodname~training across various tasks.}
    \label{tab:output_examples_2}
\end{table}
\begin{table}[htbp]
\small
    \centering
    \begin{tabular}{p{5.5cm}|p{3cm}|p{2.5cm}}
    \toprule
    \makecell[c]{\textbf{Question}} & \makecell[c]{\textbf{Before training}} & \makecell[c]{\textbf{After training}}  \\   \toprule

     Alice, Bob, and Claire are friends and avid readers who occasionally trade books. At the start of the semester, they each buy one new book: Alice gets Ulysses, Bob gets Frankenstein, and Claire gets Catch-22. As the semester proceeds, they start trading around the new books. First, Bob and Claire swap books. Then, Claire and Alice swap books. Finally, Alice and Bob swap books. At the end of the semester, Bob has Options: (A) Ulysses (B) Frankenstein (C) Catch-22. Let's think step by step. & Frankenstein = 2/ 3 of bob's books then bob gets a second book they trade around books bob eventually gets Ulysses. So, the answer is (A). & Frankenstein at the end of the semester. So the final answer is (B). \\ \midrule
    Alice, Bob, and Claire are dancers at a square dance. At the start of a song, they each have a partner: Alice is dancing with Izzi, Bob is dancing with Melissa, and Claire is dancing with Lola. Throughout the song, the dancers often trade partners. First, Claire and Alice switch partners. Then, Bob and Claire switch partners. Finally, Alice and Bob switch partners. At the end of the dance, Alice is dancing with Options: (A) Izzi (B) Melissa (C) Lola. Let's think step by step. &  At the start of the song, everyone has a partner, hence Alice starts dancing with Izzi; bob dances with Melissa; Claire dances with Lola throughout the song, they usually trade partners. So, at the end of the dance, Alice is dancing with Izzi; at the start, bob is dancing with Melissa; at the end, Claire is dancing with Lola. So, the answer is (C). & Izzi. So the final answer is (A). \\ \midrule

    Audrie tells the truth. Ryan says Audrie lies. Rashida says Ryan lies. Fletcher says Rashida tells the truth. Shaunda says Fletcher tells the truth. Does Shaunda tell the truth? Let's think step by step. & If Fletcher tells the truth, the third party does not have to believe his words in this scenario, Shaunda does not have to believe his statement. Therefore, the answer is No. & Shaunda says Fletcher tells the truth Shaunda is telling the truth because fletcher tells the truth. So the answer is Yes. \\ 
     
    \bottomrule
    \end{tabular}
    \vspace{0.5em}
    \caption{Model output examples on Bigbench. The table presents the output generated by the model before and after \methodname~training across various tasks.}
    \label{tab:output_examples_3}
\end{table}

\begin{table}[htbp]
\small
    \centering
    \begin{tabular}{p{5.5cm}|p{3cm}|p{2.5cm}}
    \toprule
    \makecell[c]{\textbf{Question}} & \makecell[c]{\textbf{Before training}} & \makecell[c]{\textbf{After training}}  \\   \toprule

     Here is a table where the first line is a header and each subsequent line is a penguin: name, age, height (cm), weight (kg) Louis, 7, 50, 11 Bernard, 5, 80, 13 Vincent, 9, 60, 11 Gwen, 8, 70, 15 For example: the age of Louis is 7, the weight of Gwen is 15 kg, the height of Bernard is 80 cm. And here is a similar table, but listing giraffes: name, age, height (cm), weight (kg) Jody, 5, 430, 620 Gladys, 10, 420, 590 Marian, 2, 310, 410 Donna, 9, 440, 650 How many penguins are there in the tables? Options: (A) 1 (B) 2 (C) 3 (D) 4 (E) 5. Let's think step by step. & Number of giraffes is twice as many as the number of penguins in the first table. So, there are 2 tables. The answer: (C). & There are 4 penguins in the first table and 4 giraffes in the second table. The answer: (D). \\ \midrule
    Here is a table where the first line is a header and each subsequent line is a penguin: name, age, height (cm), weight (kg) Louis, 7, 50, 11 Bernard, 5, 80, 13 Vincent, 9, 60, 11 Gwen, 8, 70, 15 For example: the age of Louis is 7, the weight of Gwen is 15 kg, the height of Bernard is 80 cm. Which penguin is one year younger than Vincent? Options: (A) Louis (B) Bernard (C) Vincent (D) Gwen (E) James. Let's think step by step. &  Louis = 7*11 - 9*60 = 1 year younger than Vincent. The answer: (A). & Louis penguin Louis, 7, 50, 11 Bernard, 5, 80, 13 Vincent, 9, 60, 11 Gwen, 8, 70, 15. So the answer is (D). \\ \midrule

    The following paragraphs each describe a set of five objects arranged in a fixed order. The statements are logically consistent within each paragraph. On a branch, there are five birds: a falcon, a raven, a cardinal, a hawk, and a crow. The cardinal is the second from the right. The hawk is to the right of the falcon. The raven is the second from the left. The crow is the rightmost. Options: (A) The falcon is the rightmost (B) The raven is the rightmost (C) The cardinal is the rightmost (D) The hawk is the rightmost (E) The crow is the rightmost. Let's think step by step. & The crow belongs to the rightmost group. The answer: (C). & The crow is the rightmost. The answer: (E). \\ 

    \bottomrule
    \end{tabular}
    \vspace{0.5em}
    \caption{Model Output Examples. The table presents the output generated by the model before and after \methodname~training across various tasks.}
    \label{tab:output_examples_4}
\end{table}

\clearpage
\section{Algorithm Description}

\begin{algorithm}
	\renewcommand{\algorithmicrequire}{\textbf{Input:}}
	\renewcommand{\algorithmicensure}{\textbf{Output:}}
	\caption{Self-Improvement by Reinforcement Learning Contemplation}
	\label{alg1}
	\begin{algorithmic}[1]
		\STATE \textbf{Input}: a pre-trained LLM $\gM$, a pre-trained LLM for self-evaluation $\gM^*=\gM$, an unlabelled dataset $\gD^{train}=\{q_i\}_{i=1}^{|\gD|}$.

		\REPEAT
		\STATE Sample questions $\{q_i\}$ from dataset $D^{train}$.
		\STATE Use $\gM$ to sample answers $\{o_i\}$ to the questions, with stochastic output of LLM.
		\STATE Obtain reward $R(q,o)$ according to self-evaluation results (Eq. \ref{eq:reward}).
  	\STATE Update $\gM$ with RL algorithm.

		\UNTIL training completion.
	\end{algorithmic}  
 \label{algorithm}
\end{algorithm}

\section{Societal Impact}
The introduction of \methodname~offers a promising solution to the challenges associated with fine-tuning large language models (LLMs) using external labels. This approach has the potential to reduce the cost and time required for supervision, which can increase access to NLP technologies for individuals and organizations with limited resources. Additionally, the increased accuracy and effectiveness of LLMs through the use of \methodname~has implications for a wide range of NLP tasks. This technology can improve NLP in areas such as machine translation, reasoning problems, and text generation. As a result, \methodname~has the potential to influence diverse industries such as healthcare, finance, and education, where accurate NLP model is essential for effective decision-making. \methodname~represents an exciting advancement in the field of NLP that could impact society, provided that the self-improvement of the LLMs remains aligned with human values.

\end{document}